\UseRawInputEncoding
\documentclass[a4paper,fleqn]{cas-dc}

\usepackage[numbers]{natbib}
\usepackage{booktabs}
\usepackage{multirow}
\usepackage{amsmath}
\usepackage{algorithm}
\usepackage{algpseudocode}
\usepackage{graphicx}
\usepackage[dvipsnames]{xcolor}
\usepackage{xurl}
\usepackage{hyperref}
\usepackage[nameinlink,noabbrev]{cleveref}
\usepackage{placeins}
\usepackage{caption}
\usepackage{balance}
\usepackage[normalem]{ulem}
\usepackage[most]{tcolorbox} % for section-wide highlighting

\providecommand{\printorcid}{}
\providecommand{\printfacebook}{}
\providecommand{\printtwitter}{}
\providecommand{\printlinkedin}{}
\providecommand{\printgplus}{}

\providecommand{\printemails}{}
\providecommand{\printurls}{}
\begin{document}
\let\WriteBookmarks\relax
\def\floatpagepagefraction{1}
\def\textpagefraction{.001}

% ────────────────────────────── Front matter ──────────────────────────────

{\shorttitle{R4Tun: LLM-guided adaptive tunnel segmentation}
\shortauthors{X.\ Tao et~al.}

\title[mode=title]{R4Tun: LLM-guided adaptive segmental tunnel lining segmentation in point clouds}

\author[1]{Xinghui Tao}
\credit{Conceptualization, Methodology, Software, Writing -- original draft}

\author[2]{Zehao Ye}

\author[1]{Guangming Wang}
\cormark[1]
\ead{gw462@cam.ac.uk}
\credit{Supervision, Writing -- review \& editing}

\author[2]{Jelena Nini\'{c}}
\credit{Supervision, Writing -- review \& editing}

\author[1]{Brian Sheil}
\credit{Supervision, Funding acquisition, Project administration}

\cortext[cor1]{Corresponding author}

\affiliation[1]{organization={Construction Engineering, University of Cambridge},
    addressline={Trumpington Street},
    city={Cambridge},
    postcode={CB2 1PZ},
    country={United Kingdom}}

\affiliation[2]{organization={Department of Engineering, Durham University},
    addressline={Stockton Road},
    city={Durham},
    postcode={DH1 3LE},
    country={United Kingdom}}

% ─── Abstract ─────────────────────────────────────────────────────────────

\begin{abstract}
Automated inspection of segmental tunnel linings requires adaptive segmentation from 3D point clouds, yet expert-tuned pipelines often degrade when tunnel conditions vary. This paper presents R4Tun, a large language model (LLM)-driven adaptation framework that extends an expert-designed pipeline (SAM4Tun) with bounded parameter tuning informed by structured context: memory ($m$), state ($s$), and knowledge ($k$). Evaluated on 30 selected Seg2Tunnel subsets (13 regular, 17 complex) across three LLMs, the full $m+s+k$ design raised mean Intersection-over-Union (mIoU) from 0.18 to 0.43--0.48 and overall accuracy (OA) from 0.42 to 0.59--0.65 relative to the static SAM4Tun baseline, with the near-reference regular (staggered) subsets reaching mIoU 0.784--0.796 across LLMs. Across 270 (30 tunnels $\times$ 3 different LLMs $\times$ 3 context settings) runs, the LLMs showed similar parameter-adjustment trends (with overlapping 95\% CIs on mean gains) and consistently adjusted a shared set of critical parameters. These results support R4Tun as a controlled, label-free, cross-LLM adaptation mechanism in the tested SAM4Tun--Seg2Tunnel setting, demonstrating consistent accuracy gains; we position R4Tun as a mechanism contribution rather than a deployable final-inspection system, in which each bounded parameter change is auditable via logged rationales.
\end{abstract}

% ─── Highlights ───────────────────────────────────────────────────────────

\begin{highlights}
\item Propose R4Tun for label-free adaptation of tunnel lining segmentation pipelines.
\item Integrate memory, state, and knowledge for multi-agent parameter tuning.
\item Achieve mIoU of 0.784--0.796 on near-reference tunnel subsets across different LLMs.
\item Raise overall mIoU from 0.18 to 0.43--0.48 without labels or retraining.
\item Enable controlled cross-LLM adaptation without retraining the expert pipeline.
\end{highlights}

% ─── Keywords ─────────────────────────────────────────────────────────────

\date{\today}

\maketitle

\smallskip
% ─────────────────────────────────────────────────────────────────────────

% ══════════════════════════════════════════════════════════════════════════
\section{Introduction}\label{sec:intro}
% ══════════════════════════════════════════════════════════════════════════

Automated inspection of segmental tunnel linings is required to support structural health assessment and long-term operational safety, given the scale, frequency, and access constraints of modern tunnel networks~\cite{attard2018}. In practice, engineers routinely encounter projects in which tunnel geometry, lining properties, and data acquisition settings vary~\cite{weidner2024generalized,yue2024damage}. Furthermore, although modern laser scanning enables high-fidelity tunnel capture, the resulting measurements often contain mixed structural elements, occlusions, and noise, making direct extraction of lining components challenging~\cite{huang2021bim,sjolander2023towards}. Consequently, segmentation methods need to be adaptive enough to handle this complexity and variability~\cite{montero2015, strauss2020sensing}.

\begin{figure*}[t]
\centering
\includegraphics[width=\textwidth]{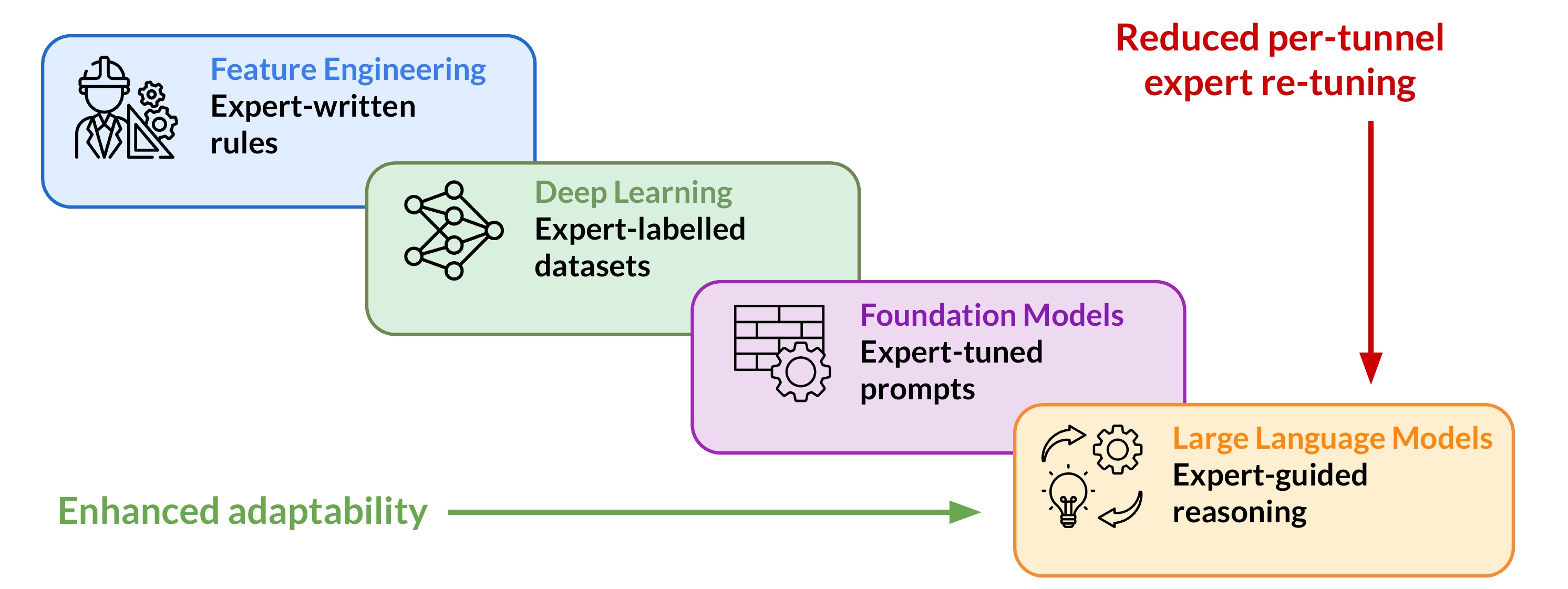}
\caption{Segmentation approaches arranged by their adaptation mechanisms. Expert-guided, LLM-driven methods can extend foundation-model pipelines by adding parameter adaptation with structured context and encoded expert knowledge. In this work, we test one such mechanism (R4Tun) while keeping the underlying SAM4Tun pipeline fixed.}
\label{fig:paradigms}
\end{figure*}

% Tunnel point-cloud segmentation methods in this study are grouped into three broad categories
Existing tunnel point-cloud segmentation methods are grouped into three broad categories: feature engineering, deep learning and foundation-model pipelines
(see Fig.~\ref{fig:paradigms} and Section~\ref{sec:related}). Feature engineering rules are interpretable but lack robustness under changed conditions~\citep{weidner2024generalized}. Supervised deep-learning models generalise through data but require large labelled datasets and periodic retraining, while lacking auditability~\citep{duan2023high, cha2024deep, su2024review}. Foundation-model pipelines bypass annotation by using models pre-trained on large datasets~\citep{bommasani2021foundation, kirillov2023sam}, but remain sensitive to expert-tuned processing parameters. Existing tunnel-segmentation approaches have not yet demonstrated, in a single workflow, both robust adaptation to changed conditions and an auditable adaptation process. Among foundation-model pipelines, SAM4Tun~\citep{ye2025sam4tun} combines point-cloud preprocessing with SAM-based prompting~\citep{kirillov2023sam} and has shown strong performance under expert tuning. However, the pipeline is highly sensitive to its many processing parameters. When tunnel geometry or scanning conditions depart from the tuning reference, performance can degrade, and identifying which parameters need adjustment can require repeated expert intervention.

Given that recent LLMs can follow multi-step instructions over structured inputs, parameter adaptation can be formulated as a diagnostic task in which the model receives context and proposes bounded adjustments. LLMs have been applied to engineering workflows including chemical synthesis planning~\citep{bran2024chemcrow}, multi-agent coordination for complex project tasks~\citep{qian2024chatdev, hong2024metagpt, chen2024comm}, and manufacturing decision support~\citep{garcia2024manufacturing}. However, these studies do not address parameter adaptation in infrastructure inspection pipelines. In this study, we propose R4Tun, an LLM-guided adaptation system that extends the SAM4Tun pipeline by adjusting stage parameters in response to changing tunnel conditions. The framework provides each agent with structured context $m+s+k$ (memory, state, and knowledge), from which it produces bounded parameter updates via stepwise prompting. We evaluate whether this contextual information improves segmentation adaptability across both regular and complex tunnels, and whether the resulting adaptation behaviour is consistent across LLMs. To isolate the contribution of the LLM beyond deterministic tuning, we also benchmark R4Tun against a rule-based adaptation derived from the same per-stage knowledge documents.

We position R4Tun as a mechanism contribution rather than a deployable final-inspection system: its strength lies in integrating LLM-driven bounded parameter adaptation with an expert-designed pipeline, with each change auditable through logged rationales. Absolute segmentation accuracy remains bounded by the open-source SAM4Tun ceiling and by the single expert reference used here, as discussed in \Cref{sec:error-analysis}.

This paper makes the following contributions:
\begin{enumerate}
\item A framework design in which LLM agents adapt the parameters of the SAM4Tun pipeline using structured context $m+s+k$: memory of the reference configuration, state from intermediate pipeline outputs, and a knowledge document of tunnel category-specific configuration shared across all tunnels.
\item A controlled and auditable adaptation mechanism rather than a deployable final-inspection system, in which each bounded parameter update is accompanied by a logged rationale for post-hoc expert review.
\item Empirical evidence from cumulative ablation across 30 tunnels suggesting that intermediate pipeline state is an important contributor to adaptation, while the knowledge component adds a smaller increment concentrated on the complex category.
\item Cross-LLM evidence showing that three different LLMs (Opus-4.6, GPT-5.4, Gemini-3-Flash) produce overlapping effect ranges and adjust a consistent subset of critical parameters under the same prompt structure.
\item Evidence that deterministic, rule-based adaptation explains part, but not all, of the observed mIoU increase, which is consistent with an additional contribution from the LLM-based adaptation process in the tested setting.
\end{enumerate}

The rest of this paper is organised as follows. Section~\ref{sec:related} reviews related work. Section~\ref{sec:methods} presents methodology, including the R4Tun architecture, dataset, experimental design, and evaluation. Section~\ref{sec:results} reports results. Section~\ref{sec:discussion} discusses findings, implications, and limitations. Section~\ref{sec:conclusions} concludes.

% ══════════════════════════════════════════════════════════════════════════
\section{Related work}\label{sec:related}
% ══════════════════════════════════════════════════════════════════════════
This study situates R4Tun within tunnel point-cloud segmentation by contrasting (i) feature-engineered methods and supervised deep learning, (ii) foundation-model pipelines that reduce annotation but remain parameter-sensitive, and (iii) LLM reasoning as a way to produce auditable, bounded per-tunnel parameter updates.

\begin{figure*}[htp]
\centering
\includegraphics[width=0.8\textwidth]{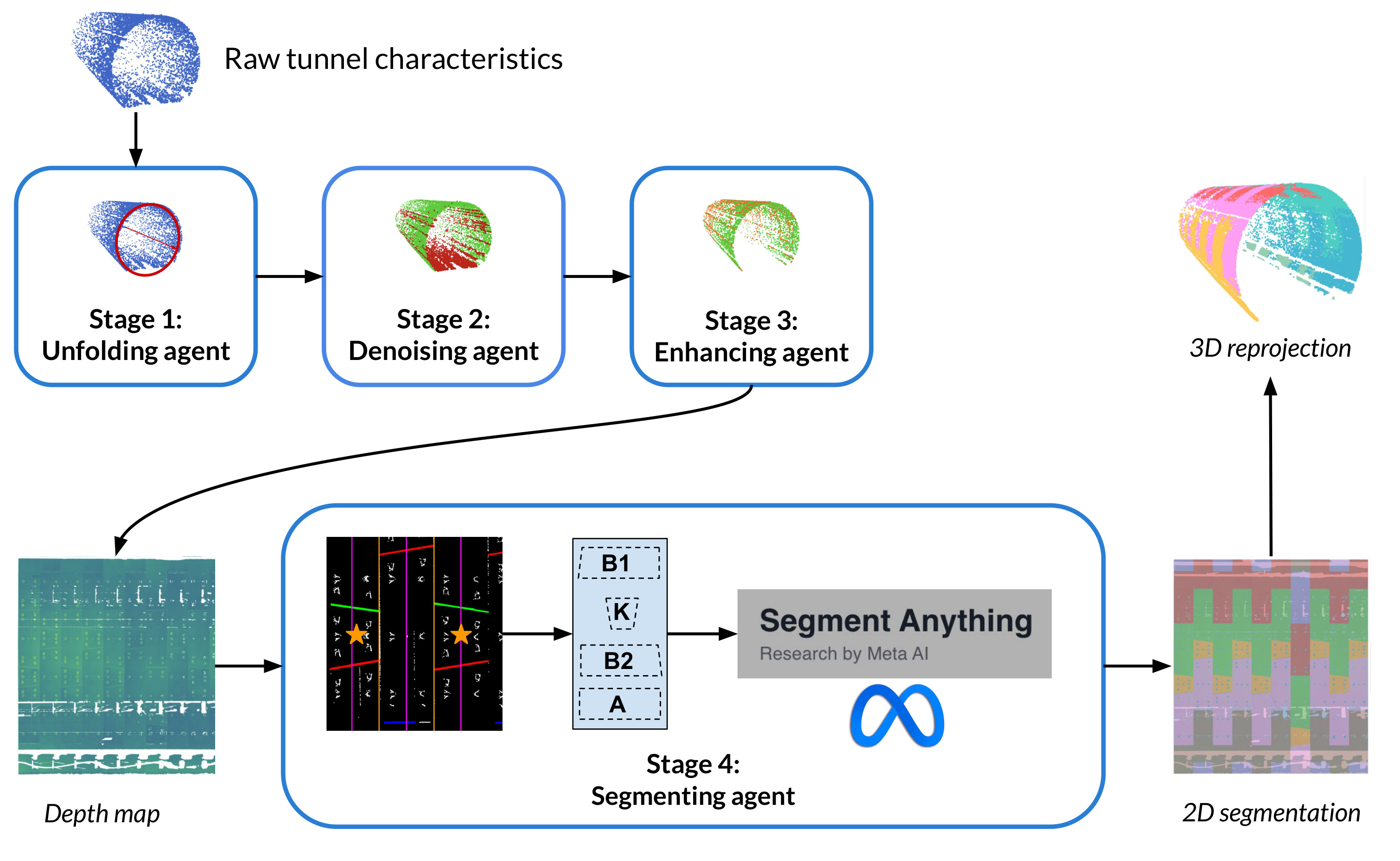}
\caption{The R4Tun multi-agent architecture for tunnel segmentation. Four blue-boxed stage agents (Stage~1: Unfolding, Stage~2: Denoising, Stage~3: Enhancing, and Stage~4: Segmenting) adapt their stage-specific parameters via LLM reasoning, passing intermediate outputs forward to produce the final segmentation. Arrows indicate the workflow direction from the raw point cloud through intermediate representations to 2D segmentation and 3D reprojection.}
\label{fig:architecture}
\end{figure*}

\subsection{Feature engineering versus deep learning}

Tunnel point-cloud segmentation has advanced through two generations of methods, which trade off between auditability and adaptability. Feature-engineering approaches encode domain knowledge into deterministic geometric rules: centreline fitting via RANSAC~\cite{fischler1981ransac}, density-based surface clustering~\cite{ester1996dbscan}, and point-sampled surface processing (e.g., simplification and resampling)~\cite{pauly2002}. Those methods remain widely used in safety-critical inspection because their logic is explicit and auditable~\cite{weidner2024generalized}. However, each rule embeds assumptions about tunnel geometry and point-cloud characteristics (diameter, ring spacing, joint pattern, point density), and changing any of these requires manual reconfiguration by a domain expert. Supervised deep-learning methods, such as PointNet++~\cite{qi2017pointnet}, RandLA-Net~\cite{hu2020randla}, Mask3D~\cite{schult2023mask3d}, and TD3D~\cite{kolodiazhnyi2024td3d}, learn features directly from annotated point clouds and can generalise across similar conditions present in the training set. Recent work has also extended tunnel and infrastructure understanding with attention mechanisms: for 3D point clouds, UnrollingNet~\cite{zhang2022unrollingnet} unfolds the cylindrical surface into a structured 2D representation for attention-based semantic segmentation, and SerialFormer~\cite{wang2026serialformer} uses transformer-style attention to capture long-range context for point-cloud semantic segmentation; for 2D inspection imagery, TransUNet~\cite{li2026transunet} leverages self-attention within a U-Net-style encoder--decoder for tunnel defect segmentation, and ECA-YOLO~\cite{liang2026ecayolo} augments a {YOLO} detector with efficient channel attention and tiling to improve multi-defect detection. Yet supervised
deep-learning models require large labelled tunnel datasets that remain scarce~\cite{su2024review}, demand retraining for new tunnel conditions, and provide no mechanism for an engineer to trace or override a segmentation decision~\cite{cha2024deep}.

Therefore, for inspection operators, deploying to a new tunnel type requires either re-engineering rules, collecting new training data, or accepting degraded performance without a systematic diagnostic mechanism.

\subsection{Foundation-model pipelines}

To the best of our knowledge, direct 3D point-cloud segmentation of infrastructure at tunnel scene scale remains an open challenge for current foundation models. Existing 3D foundation models target common object recognition on curated datasets and do not transfer to large, noisy tunnel point clouds~\cite{camuffo2022recent, liu2023openshape, guo2023pointbind, xu2024pointllm}. As a result, practical tunnel pipelines often project 3D data into 2D representations to use mature vision foundation models pre-trained on large-scale image datasets, thereby bypassing the annotation bottleneck~\cite{caron2021dino, franceschelli2024training}. SAM~\cite{kirillov2023sam} performs class-agnostic segmentation from spatial prompts without task-specific training, and extensions such as SAM2~\cite{ravi2024sam2} for temporal consistency and SEEM~\cite{zou2023segment} for natural-language-guided segmentation are broadening prompt-based segmentation further. These models have been adopted in infrastructure contexts including crack detection~\cite{ge2024cracksam,ye2024sam} and Scan-to-BIM workflows~\cite{wang2024omni, pan2024zero,ye2026maintenance}.

For tunnel linings, SAM4Tun~\cite{ye2025sam4tun} combines geometric preprocessing (unfolding, denoising, enhancing) with foundation-model 2D segmentation, achieving strong segmentation without training data. However, this pipeline depends on numerous stage-specific parameters that encode assumptions about the reference tunnel's diameter, ring length, point density, and joint patterns. When a new tunnel departs from the reference, these parameters become misspecified and the pipeline can degrade without indicating which parameters fail or why. This sensitivity is not unique to SAM4Tun: most multi-stage pipelines that chain domain-specific preprocessing with a foundation model tend to inherit it. While foundation models have reduced reliance on labelled data, they have made the pipeline's dependence on hand-tuned parameters more explicit, without removing the need for expert parameter tuning.

\subsection{LLM reasoning}

Recent LLM advances have introduced explicit reasoning capabilities relevant to the parameter-adaptation problem. Reasoning-oriented training~\cite{ouyang2022training, deepseek2025analysis, chua2025deepseek, xiang2025metacot, openai2025o3, openai2025reasoningbestpractices} produces models capable of decomposing complex problems and performing prompted consistency checks. Chain-of-thought (CoT) reasoning~\cite{wei2023cot, kojima2023zeroshot} enables step-by-step logical traces. Multi-agent architectures coordinate specialised roles through shared context~\cite{hong2024metagpt, qian2024chatdev, chen2024comm, zhang2023igniting, zhang2024chain}, and context engineering influences reasoning quality through the careful design of information provided to models~\cite{xu2024retrieval, mei2025survey, anthropic2025context}.

In tunnelling segmentation, however, no prior work has applied LLM-based reasoning to a challenge in expert-designed pipelines: re-tuning a parameter-sensitive system to new conditions while preserving the engineer's ability to inspect and override each decision. We therefore introduce an LLM-based reasoning framework that retunes parameter-sensitive pipelines per tunnel using expert-guided context and a logged rationale for each proposed change. Meanwhile, whether LLM reasoning produces gains beyond what hand-coded rules can deliver from the same knowledge text remains an open question. Our experimental design therefore tests this directly by including a deterministic rule-based adaptation of those documents as a non-LLM control.

% ══════════════════════════════════════════════════════════════════════════
\section{Methodology}\label{sec:methods}
% ══════════════════════════════════════════════════════════════════════════

Our methodology defines the tunnel-lining segmentation task, summarises the fixed SAM4Tun pipeline, and specifies the R4Tun adaptation layer (structured context \mbox{$m+s+k$} and bounded parameter updates), followed by the dataset, baselines, ablations, metrics, and sensitivity analysis.

\subsection{Task definition}\label{sec:problem-def}

The task is semantic segmentation of segmental tunnel linings from terrestrial laser scanning (TLS) point clouds. Given a raw point cloud of a tunnel section, the goal is to assign each point a structural label: background, key segment~(K), adjacent segments (B), and standard segments (A), with an additional A4 class for 7-segment complex tunnels. The unit of analysis is a tunnel subset spanning multiple rings selected from the Seg2Tunnel benchmark. 

A key limitation is that the fixed configuration of the open-source SAM4Tun achieves an mIoU of 0.27 on regular tunnels, but only 0.04 on complex tunnels whose geometry departs from the tuning reference. As noted above, the pipeline provides no built-in signal indicating which parameters become misspecified.

\subsection{R4Tun}\label{sec:method}

\begin{figure*}[htp]
\centering
\includegraphics[width=0.8\textwidth]{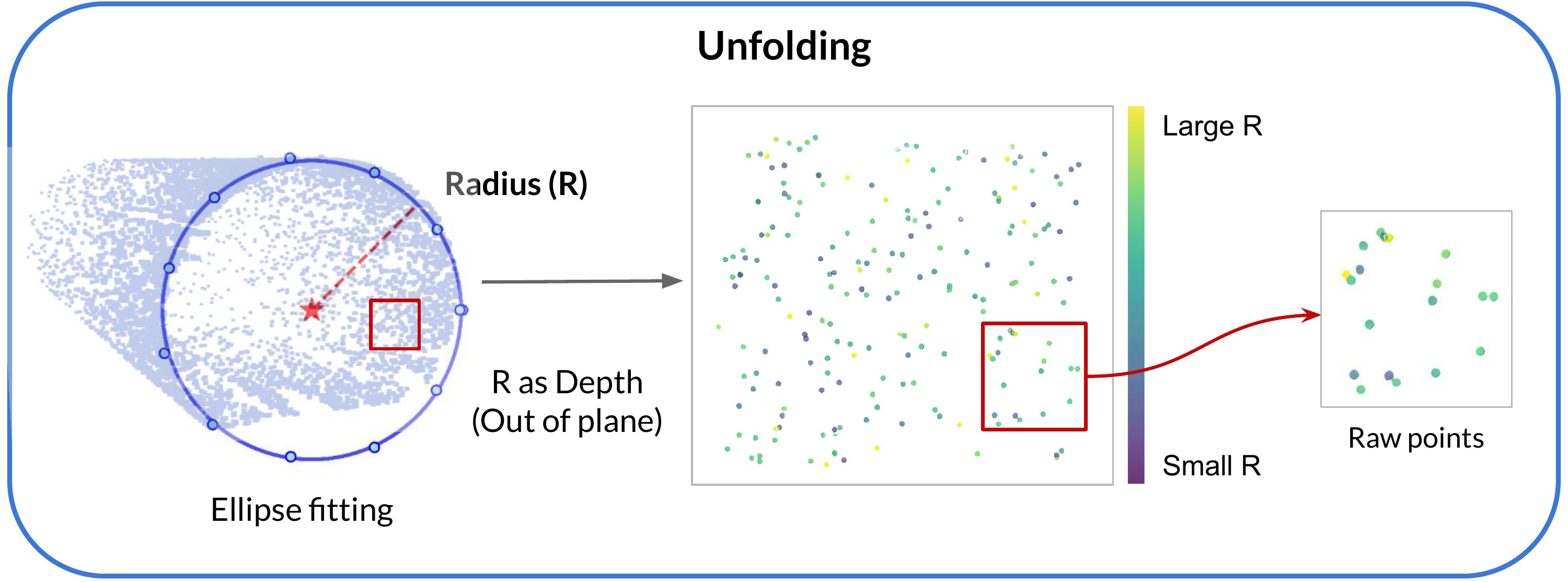}
\caption{Stage~1: Unfolding. Left: fitted tunnel cross-section used to determine the reference centre and radius for cylindrical unfolding. Centre: 2D panoramic depth map derived from the 3D point cloud. Right: zoomed region of the unfolded map showing radial-distance encoding.}
\label{fig:stage-u}
\end{figure*}

R4Tun supplements the SAM4Tun pipeline~\cite{ye2025sam4tun} with an LLM-guided adaptation layer that adjusts parameters per tunnel without modifying the underlying algorithms (Fig.~\ref{fig:architecture}). Given a new tunnel point cloud, a characteriser first extracts raw geometric properties (diameter, density, coordinate ranges; full field list in Appendix~\ref{app:chars}). Each stage (Stage~1: Unfolding, Stage~2: Denoising, Stage~3: Enhancing, and Stage~4: Segmenting) has a dedicated LLM agent that adapts its parameters. In Stage~4, the adapted prompts are passed to SAM, and the resulting 2D masks are reprojected onto the 3D point cloud without further LLM intervention.

\textbf{Stage~1: Unfolding} (Fig.~\ref{fig:stage-u}) establishes a cylindrical coordinate system for the tunnel. Similar cylindrical unfolding ideas have been explored for tunnel point-cloud semantic segmentation (e.g., UnrollingNet~\cite{zhang2022unrollingnet}). It slices the point cloud at regular longitudinal intervals, fits an ellipse to each cross-section via RANSAC~\cite{fischler1981ransac}, and selects the model with the highest inlier support. The per-slice ellipse centres are interpolated into a smooth centreline that defines the cylindrical frame~$(r, \theta, h)$. The 3D tunnel surface is then projected into a 2D panoramic image in which each pixel encodes the radial distance to the centreline, producing a stable depth representation for downstream filtering and segmentation.

\begin{figure*}[htp]
\centering
\includegraphics[width=0.8\textwidth]{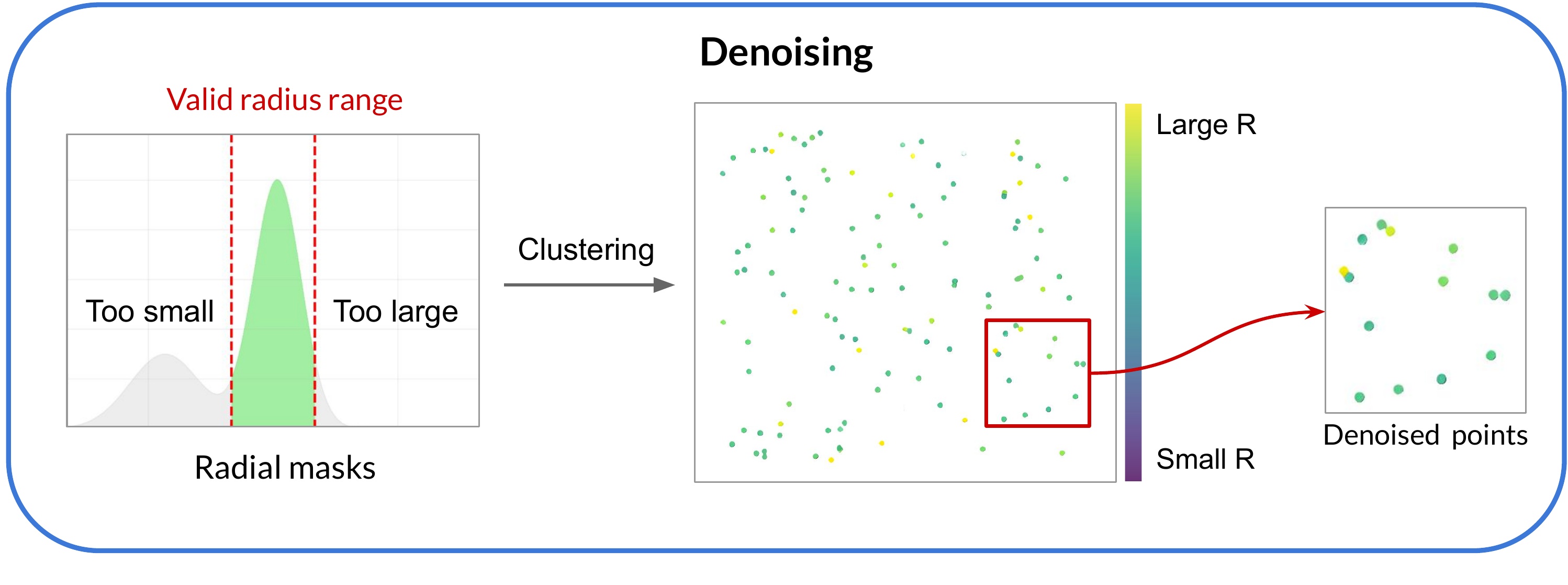}
\caption{Stage~2: Denoising. Left: radial-distance distribution used to filter points outside the expected tunnel lining surface. Centre: unfolded 2D depth map after filtering, showing the cleaned point distribution. Right: zoomed region of the denoised map.}
\label{fig:stage-d}
\end{figure*}

\textbf{Stage~2: Denoising} (Fig.~\ref{fig:stage-d}) removes non-structural artefacts (rails, cables, and scattered points) that would otherwise interfere with segmentation. The algorithm applies grid-based radial-density filtering inspired by the DBSCAN clustering principle~\cite{ester1996dbscan}, grouping points by local density in cylindrical coordinates. Dense, connected regions are retained as tunnel surfaces while isolated points are rejected as noise. A pair of radial masks~($R_\mathrm{low}$, $R_\mathrm{high}$) constrains filtering to the expected tunnel radius, preserving joint boundaries and ring edges while discarding outliers beyond the lining surface. The result is a clean, continuous surface representation suitable for curvature analysis and interpolation in the next stage. 

\begin{figure*}[htp]
\centering
\includegraphics[width=0.8\textwidth]{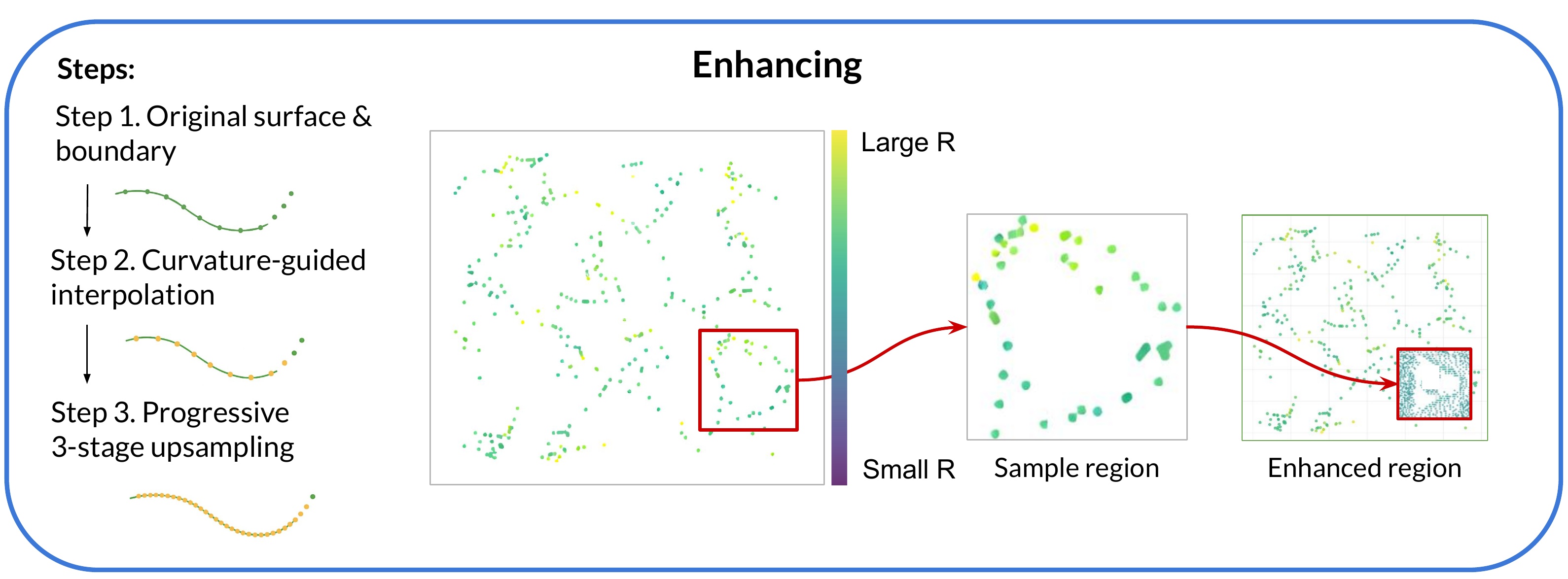}
\caption{Stage~3: Enhancing. Left: curvature-guided surface and boundary interpolation along the tunnel lining, showing original sampling, curvature-guided insertion, and progressive three-stage upsampling (shown as three sequential upsampling steps). Centre: unfolded 2D depth map after enhancement, where inserted points improve geometric continuity; the red box highlights a local region of interest. Right: zoomed views of the highlighted region, showing added points and pixel-level interpolation.}
\label{fig:stage-e}
\end{figure*}

\textbf{Stage~3: Enhancing} (Fig.~\ref{fig:stage-e}) improves geometric continuity and surface completeness before projection into the 2D depth map. Local curvature is estimated on neighbourhood points to guide point insertion: new points are placed between neighbours whose curvature difference is small (smooth panel regions), while high-curvature areas (joint boundaries and ring edges) are preserved intact. A three-stage progressive upsampling scheme inserts additional points at progressively finer resolutions to maintain surface continuity without blurring structural boundaries. Pixel-level interpolation then refines outlier points at joint locations. The enhanced surface is projected into a panoramic depth map that balances smooth panel regions with sharply defined joint boundaries. 

\begin{figure*}[htp]
\centering
\includegraphics[width=0.8\textwidth]{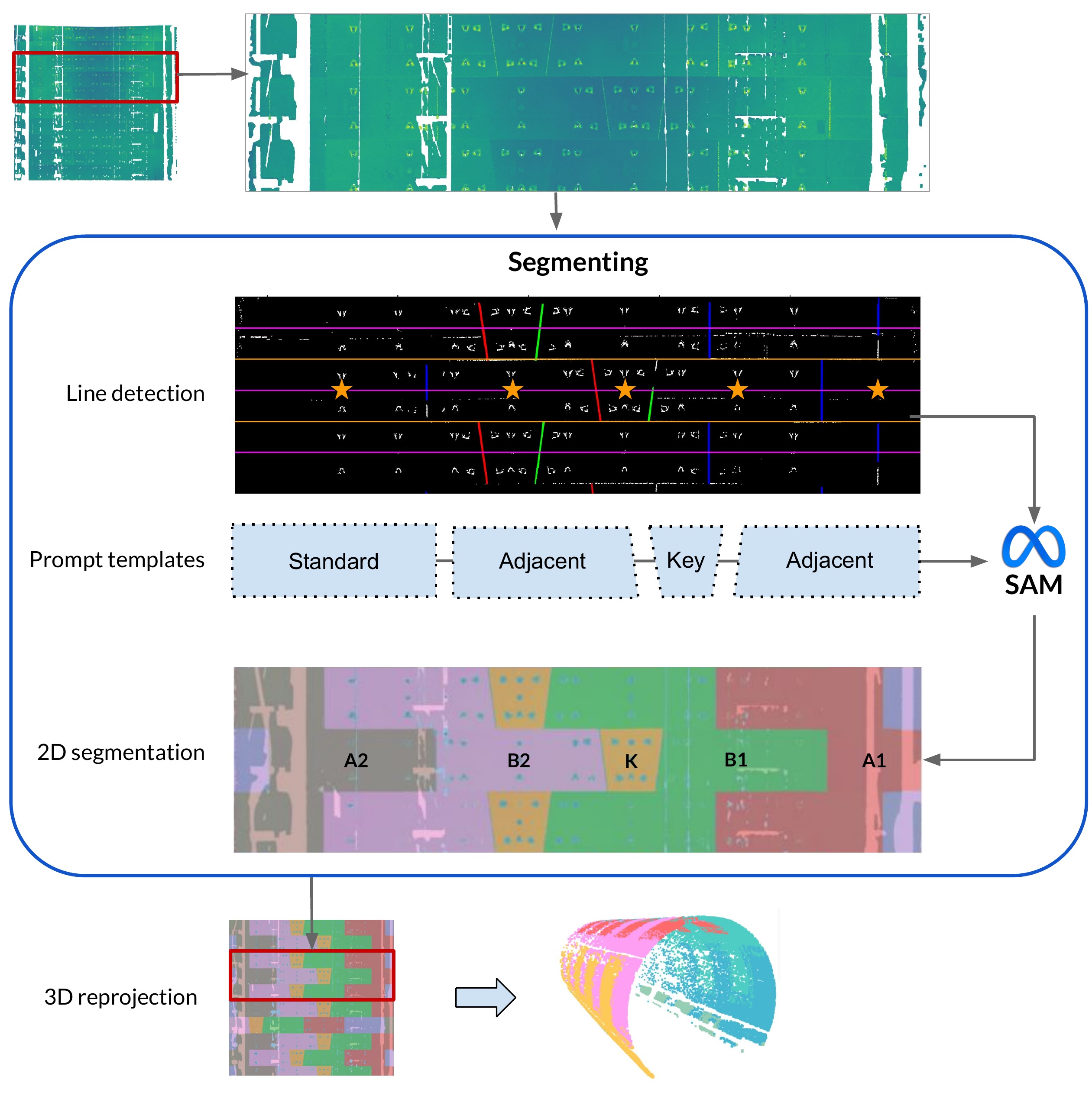}
\caption{Stage~4: Segmenting. The segmenting stage applies Hough-transform line detection~\cite{duda1972hough} to identify ring boundaries, then constructs template-based prompts and feeds them to SAM~\cite{kirillov2023sam}. SAM produces 2D segment masks that are reprojected into 3D. K denotes the key segment, B denotes adjacent segments, and A denotes standard segments around the ring.}
\label{fig:stage-p}
\end{figure*}

\textbf{Stage~4: Segmenting} (Fig.~\ref{fig:stage-p}) combines boundary detection and SAM segmentation into a single workflow. First, a Hough-transform-based detector~\cite{duda1972hough} extracts linear features from the enhanced depth map that correspond to ring joint boundaries. Detected lines are filtered by orientation (oblique, horizontal, vertical) with separate Hough thresholds and merged when closer than a distance tolerance. The confirmed boundaries are then used to construct template-based prompts for key, adjacent, and standard segments, which are passed to SAM~\cite{kirillov2023sam}. SAM produces 2D segment masks on the panoramic image, which are reprojected into 3D point labels via the stored pixel-to-point mapping from Stage~1.

The pipeline contains 81 tunable processing parameters, spanning geometric defaults, filtering and interpolation settings, and boundary-detection thresholds. All parameters were expert-tuned on a reference tunnel (mIoU\,=\,0.88); full parameter tables are provided in Appendix~\ref{app:params}. In this study, the underlying pipeline stages remain fixed across all conditions; only the parameter values change. The SAM4Tun baseline applies this single expert-tuned configuration uniformly to all 30 tunnels without per-tunnel adaptation, serving as the static baseline against which LLM-guided adaptation is measured.

\subsection{Agent design}\label{sec:agent-design}

\begin{figure*}[t]
\centering
\includegraphics[width=0.8\textwidth]{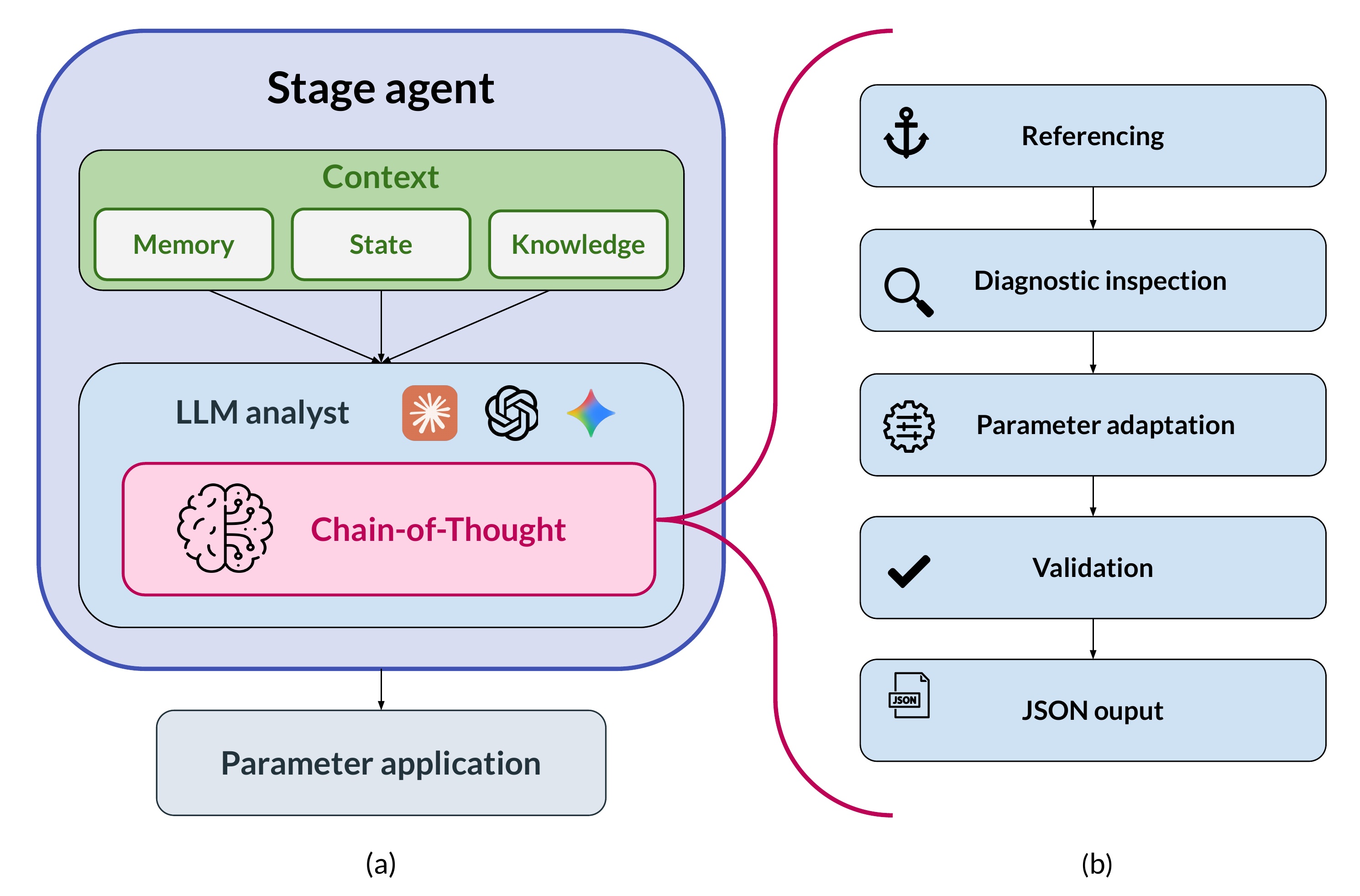}
\caption{The left column (a) shows R4Tun stage agent architecture. Each agent maintains a shared and cumulative context that informs an LLM analyst component performing CoT reasoning. The right column (b) shows the five CoT phases: (1) Referencing, (2) Diagnostic inspection, (3) Parameter adaptation, (4) Validation, and (5) JSON output.}
\label{fig:agent-design}
\end{figure*}

Each of the four stage agents is a self-contained unit comprising (i)~a \emph{context} (Section~\ref{sec:context-comp}; Fig.~\ref{fig:agent-design}a) and (ii)~a Python \textit{analyst} that constructs the full LLM prompt and parses the returned JSON. For each agent-driven stage, the analyst assembles a prompt from the current context level, calls one of the selected LLM APIs (Opus-4.6, GPT-5.4, or Gemini-3-Flash) to follow a five-step CoT protocol (Section~\ref{sec:cot}; Fig.~\ref{fig:agent-design}b): referencing, diagnostic inspection, parameter adaptation, validation, and JSON output. Each agent returns a single schema-conformant parameter JSON, which the corresponding pipeline stage executes unchanged. A characteriser plugin then extracts statistics from the stage output, updating the cumulative state available to subsequent stages. Each stage's parameters are adapted from the expert-tuned reference configuration together with this cumulative state; the stage scripts and evaluation code are identical across all ablation conditions, with only the parameter JSONs changing. Because every parameter change is logged alongside a generated textual rationale, engineers can review and, where necessary, override adaptation decisions. A worked example is provided in Appendix~\ref{app:cot}. After the final segmentation, per-point labels are evaluated against ground truth using mean Intersection over Union (mIoU) as the primary metric.

\subsubsection{Context design}\label{sec:context-comp}

Each stage agent receives structured context comprising three components: memory, state and knowledge (Fig.~\ref{fig:agent-design}a). A worked example of the three components for the denoising agent is given in Appendix~\ref{app:context}.

\textbf{Memory} (Appendix~\ref{app:context-memory}) stores the reference tunnel's characteristics (a JSON file containing geometry, point density, coordinate ranges, nearest-neighbour distances) alongside the expert-tuned reference parameters. The agent compares the current tunnel's characteristics against this reference to quantify deviation and adjust its parameters via CoT reasoning (Section~\ref{sec:cot}). Memory is static: it does not change between stages.

\textbf{State} (Appendix~\ref{app:context-state}) captures cumulative geometric and statistical properties after each pipeline stage executes. These JSON summaries are injected into subsequent agents' prompts, and the state grows as stages execute. Importantly, state alone does not prescribe parameter changes: the numeric summaries of the actual point distribution after upstream processing must be interpreted in the context of parameter semantics and tunnel conditions before they can inform parameter updates.

\textbf{Knowledge} (Appendix~\ref{app:context-knowledge}) supplies domain-specific guidance in human-readable markdown documents. Each stage has its own knowledge document covering three types of guidance: (i) cross-tunnel variation, which describes common lining layouts and structural conditions; (ii) parameter notes, which define each tunable parameter, its empirically validated range, and proven defaults; and (iii) diagnostic rules, which provide constrained guidelines for recommended parameter adjustments. Knowledge is authored once and shared across all tunnels. 

% tunnel-type taxonomy; parameter semantics and empirically validated ranges; and diagnostic rules linking characteristics to parameter adjustments. 

\subsubsection{CoT design}\label{sec:cot}

CoT reasoning provides the analytical structure for each agent's decision-making (Fig.~\ref{fig:agent-design}b). The agent is prompted to follow a five-step protocol (Algorithm~\ref{alg:cot_adaptation}). A worked example of the full five-step trace is provided in Appendix~\ref{app:cot}.

\begin{enumerate}
    \item \emph{Referencing:} The agent compares the current tunnel's characteristics against the stored reference to quantify deviation (e.g., $+39\%$ diameter increase).
    \item \emph{Diagnostic inspection:} The agent attributes the deviation to a plausible cause and identifies which specific tunable parameters are implicated. If signals conflict, the agent is instructed to prioritise \emph{State} over \emph{Memory}, as State summarises geometry after upstream processing.
    \item \emph{Parameter adaptation:} The agent proposes updates for the implicated parameters. These updates are instructed to stay within the empirical ranges defined in the \emph{Knowledge} component.
    \item \emph{Validation:} A self-correction step performs a consistency check to confirm that the proposed values satisfy logical constraints (e.g. a radius lower bound must be smaller than the corresponding upper bound). If a constraint is violated, the value is prompted to move toward to the nearest valid bound. This check-and-correction is executed as an LLM reasoning step, rather than a deterministic software routine, and therefore does not guarantee constraint satisfaction.
    \item \emph{JSON output:} The selected parameters and their reasoning trace are packaged into a single schema-conformant JSON object.
\end{enumerate}

Each stage agent reads the tunnel characteristics and the latest pipeline state to identify deviations and decide which parameters to adjust. Starting from the expert reference value, it computes an updated setting from the post-stage statistics and clips it to the empirically observed range in the knowledge document when needed. This differs from the deterministic rule baseline (Appendix~\ref{app:rules}), which assigns each tunnel to a discrete category based on raw characteristics and then applies one fixed parameter value per family. By contrast, the LLM diagnoses parameters on the fly, resolves conflicts between live state and static memory, and adapts values continuously from the observed statistics.

\noindent\textbf{Notation for Algorithm~\ref{alg:cot_adaptation}.} $C_{\text{target}}$ denotes the raw characteristics of the current tunnel, while $M$ stores the reference configuration $(C_{\text{ref}}, P_{\text{ref}})$. $S$ is the cumulative pipeline state, and $S_{\text{current}}$ is its most recent summary. $K$ is a shared knowledge document that encodes parameter semantics, per-parameter admissible bounds $B$, and cross-parameter constraints $R$. The adaptation procedure computes a geometric deviation descriptor $\Delta_{\text{geom}}$, selects an implicated parameter subset $P_{\text{implicated}}$, proposes an updated parameter map $P_{\text{proposed}}$ (indexed as $P_{\text{proposed}}[p]$), and finally returns the schema-formatted JSON object $P_{\text{adapted}}$.

\begin{algorithm}[ht]
\caption{LLM-driven Parameter Adaptation via CoT}\label{alg:cot_adaptation}
\begin{algorithmic}[1]
\Require Target tunnel raw characteristics $C_{\text{target}}$
\Require Memory $M$ (reference characteristics $C_{\text{ref}}$, reference parameters $P_{\text{ref}}$)
\Require State $S$ (cumulative pipeline outputs / intermediate summaries)
\Require Knowledge $K$ (parameter semantics, tunable bounds $B$, and cross-parameter constraints $R$)
\Ensure Adapted stage parameter JSON $P_{\text{adapted}}$
\State \textbf{Phase 1: Referencing}
\State $\Delta_{\text{geom}} \gets \text{Compare}(C_{\text{target}}, C_{\text{ref}})$ \Comment{Quantify deviation from reference}
\State \textbf{Phase 2: Diagnostic inspection}
\State $S_{\text{current}} \gets \text{ExtractLatest}(S)$ \Comment{Most recent stage statistics}
\State \textit{Reconciliation:} If $S_{\text{current}}$ conflicts with $M$, prioritise $S_{\text{current}}$
\State $P_{\text{implicated}} \gets \text{IdentifyParameters}(\Delta_{\text{geom}}, S_{\text{current}}, K)$ \Comment{Select parameters relevant to observed deviations}
\State \textbf{Phase 3: Parameter adaptation}
\State $P_{\text{proposed}} \gets \emptyset$
\For{each parameter $p \in P_{\text{implicated}}$}
    \State $P_{\text{proposed}}[p] \gets \text{Adjust}(p, \Delta_{\text{geom}}, S_{\text{current}}, K)$ \Comment{Propose new value within $B$ guided by $K$}
\EndFor
\State \textbf{Phase 4: Validation}
\For{each parameter $p \in P_{\text{proposed}}$}
    \State $P_{\text{proposed}}[p] \gets \text{SelfValidate}\big(P_{\text{proposed}}[p],\, p,\, K\big)$
    \Comment{Execute an LLM internal reasoning step conditioned on $K$ (semantics, bounds spec $B$, constraints $R$).}
\EndFor
\State \textbf{Phase 5: JSON output}
\State $P_{\text{adapted}} \gets \text{FormatAsJSON}(P_{\text{proposed}})$
\State \Return $P_{\text{adapted}}$
\end{algorithmic}
\end{algorithm}

\subsection{Experimental setup}\label{sec:exp-setup}

\subsubsection{Dataset}\label{sec:dataset}

We selected 30 tunnel subsets from the Seg2Tunnel benchmark (Table~\ref{tab:dataset}). These subsets cover the main variations in segmental tunnel geometry (diameter, ring length, segment count, key-segment layout, and point-density distribution), providing a controlled basis for evaluating adaptation across tunnel categories.

As shown in Fig.~\ref{fig:tunnel_type}, staggered and continuous tunnels are grouped as ``regular'' ($n = 13$): they share the 5.60\,m nominal diameter, 1.2\,m ring spacing, six segments per ring, and, most importantly, regular patterns of key-segment positions similar to the reference. Complex tunnels ($n = 17$) depart from that reference in multiple coupled ways: nominal inner diameter increases by 34\% (5.60\,m $\rightarrow$ 7.50\,m), ring length by 50\% (1.2\,m $\rightarrow$ 1.8\,m), and each ring adds a seventh segment with an interleaved key-segment arrangement that does not repeat ring-to-ring. In addition, off-axis scanning yields non-uniform sampling, partial circumferential coverage, and angle-dependent range and incidence. Together, these differences alter both the geometry visible in the point cloud and the semantic targets the pipeline must recover. Consequently, a fixed parameter set cannot be assumed to transfer from the regular reference; these conditions constitute the primary stress test for whether R4Tun can adapt the same algorithmic pipeline to off-design tunnel conditions. Ground-truth labels are per-point structural class annotations provided by the Seg2Tunnel benchmark; they are used for evaluation only and are not provided to the LLM or pipeline during adaptation.

% ; estimated: 5.32\,m $\rightarrow$ 7.41\,m, $+39\%$

\begin{figure*}[t]
\centering
\includegraphics[width=0.8\textwidth]{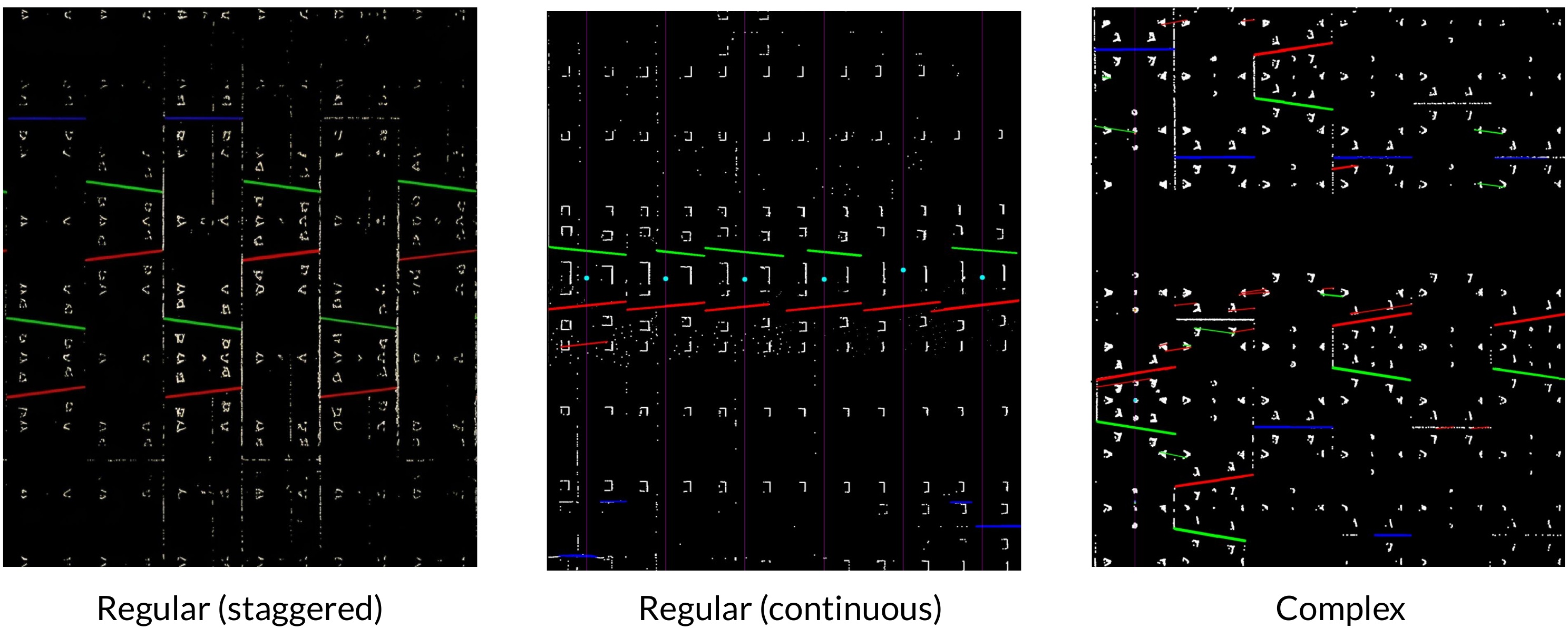}
\caption{Tunnel lining pattern categories. From left to right: regular (staggered) tunnels exhibit a repeating ring-to-ring pattern; regular (continuous) tunnels maintain a fixed segment order around each ring; complex tunnels show no consistent repeating pattern.}
\label{fig:tunnel_type}
\end{figure*}

\begin{table*}[t]
\caption{Dataset properties.}\label{tab:dataset}
\begin{tabular*}{\textwidth}{@{\extracolsep{\fill}} l l l l @{}}
\toprule
 & \multicolumn{2}{c}{Regular (T1, T2, T3)} & Complex (T4, T5) \\
\cmidrule(lr){2-3}\cmidrule(l){4-4}
Property & Staggered (T1, T2) & Continuous (T3) & \\
\midrule
Inner diameter & 5.5\,m & 5.9\,m & 7.5\,m \\
Ring length & 1.2\,m & 1.2\,m & 1.8\,m \\
Segments/ring & 6 & 6 & 7 \\
Joint type & Staggered & Continuous & Complex interleaved \\
Key-segment position & \parbox[t]{0.30\textwidth}{\raggedright A ring-to-ring repeating pattern} & \parbox[t]{0.20\textwidth}{\raggedright A fixed ring-wise order} & \parbox[t]{0.20\textwidth}{\raggedright No repeating pattern} \\
Scanning & Single-station & Multi-station & Single-station, offset \\
Eval.\ schema & 6-class & 6-class & 7-class \\
Count & 10 subsets & 3 subsets & 17 subsets \\
\bottomrule
\end{tabular*}
\end{table*}

\subsubsection{Experimental design}\label{sec:comparators}

The experiment follows a cumulative ablation design with four conditions, each adding one context component (Table~\ref{tab:ablation}). This design isolates the cumulative contribution of each component while maintaining a controlled comparison. Level~0 (SAM4Tun) is the fixed expert-tuned baseline with no LLM involvement. Levels~1--3 progressively provide the LLM with richer context, enabling the cumulative ablation to attribute improvements to specific components. A separate \mbox{\emph{Non-LLM}} condition (level~0a in Table~\mbox{\ref{tab:ablation}}) is included as the Non-LLM adaptive control, allowing the LLMs' contribution to be read directly from the gap between the \mbox{\emph{Non-LLM}} column and the LLM columns of Table~\mbox{\ref{tab:main-results}}. We codified the per-stage knowledge documents into a deterministic Python rule table as a non-LLM baseline. It maps characterisation fields (e.g., diameter, ring length, segments-per-ring, joint type, point density, and station configuration) to the 18 critical parameters adjusted by the LLMs (Table~\ref{tab:critical-params}) using explicit ``if\,\ldots\ then\,\ldots'' rules. The table and script are fixed for all 30 tunnels (regular and complex), with no evaluation-set tuning or subset-specific settings, isolating the value of LLM per-tunnel reasoning. The complete rule specification is provided in Appendix~\ref{app:rules}.

\begin{table}[ht]
\caption{Ablation conditions.}\label{tab:ablation}
\begin{tabular*}{\tblwidth}{@{} c l l >{\raggedright\arraybackslash}p{2.75cm} @{\hspace{0.7em}}@{}}
\toprule
Level & Code & Condition & What the LLM sees \\
\midrule
0 & SAM4Tun & Baseline & Default parameters \\
0a & Non-LLM & Rule table & 18 parameters set by lookup \\
1 & m & Memory & Reference characteristics + parameters \\
2 & m+s & Memory+State & + intermediate pipeline outputs \\
3 & m+s+k & \begin{tabular}[t]{@{}l@{}}Memory + State + \\ Knowledge\end{tabular} & + domain knowledge \\
\bottomrule
\end{tabular*}
\end{table}

To assess whether adaptation behaviour depends on a specific LLM, each condition was run with three different LLMs (Opus-4.6, GPT-5.4, and Gemini-3-Flash) under identical prompts. Across models, the pipeline code, evaluation scripts, and prompt structure were held constant, while the context content varied by ablation level. Other than setting temperature to 0 to minimise stochasticity (Table~\ref{tab:llm-config}), all other API settings used vendor defaults. This corresponds to 270 unique LLM-adaptation configurations: $30$ tunnels $\times$ $3$ LLMs $\times$ $3$ context settings (m, m+s, m+s+k). Each configuration was executed a second time for repeatability checking, yielding $270 \times 2 = 540$ pipeline executions, plus 30 baseline runs with fixed parameters. Headline analyses (e.g., Table~\ref{tab:main-results}) use the first execution only. All three LLMs were accessed via their respective commercial APIs (Table~\ref{tab:llm-config}). No model-specific prompt tuning was performed.

\begin{table}[ht]
\caption{LLM configuration.}\label{tab:llm-config}
\begin{tabular*}{\tblwidth}{@{} l l @{}}
\toprule
Setting & Value \\
\midrule
Models & Opus-4.6, GPT-5.4, Gemini-3-Flash \\
Max tokens & 16{,}384 \\
Temperature & 0 (override to minimise stochasticity) \\
Timeout & 300\,s per call \\
Prompt format & Markdown with JSON code \\
Failure handling & JSON extraction failure \\
\bottomrule
\end{tabular*}
\end{table}

Pipeline execution used a single NVIDIA RTX~5060 GPU. Adapted parameter files and CoT reasoning traces were logged for all runs, enabling post-hoc sensitivity analysis (Section~\ref{sec:sensitivity-method}).

\subsubsection{Evaluation metrics}\label{sec:metrics}

Segmentation quality is measured primarily by mean Intersection-over-Union (mIoU). For class $c$,
\begin{equation}
\text{IoU}_c = \frac{\text{TP}_c}{\text{TP}_c + \text{FP}_c + \text{FN}_c}, \qquad
\text{mIoU} = \frac{1}{C} \sum_{c=1}^{C} \text{IoU}_c,
\end{equation}
where $C$ is the number of segmental segments (classes) within one ring ($C=6$ for regular tunnels; $C=7$ for complex tunnels). Equation~(1) defines the per-tunnel mIoU (the mean IoU over the $C$ classes of a single tunnel); unless stated otherwise, mIoU values reported for a group of tunnels are the arithmetic mean of these per-tunnel scores across the relevant subsets. As a complementary global metric we also report overall accuracy (OA),
\begin{equation}
\text{OA} = \frac{\sum_{c=1}^{C} \text{TP}_c}{N},
\end{equation}
where $N$ is the total number of points. Because OA can be dominated by majority classes, we treat mIoU as the primary score. We also report per-class IoU to assess whether improvements are broad or concentrated in specific structural classes.

For each condition and LLM, we computed paired per-tunnel improvements as $\Delta_i = \mathrm{mIoU}_{\text{condition},i} - \mathrm{mIoU}_{\text{baseline},i}$, with $n=13$ for regular tunnels, $n=17$ for complex tunnels, and $n=30$ overall. We summarise these paired improvements using three standard statistics: (i) two-sided paired $t$-test $p$-values ($\alpha = 0.05$) for statistical significance, (ii) paired Cohen's \mbox{$d$} for effect size, and (iii) bootstrap 95\% confidence intervals (CIs) for the mean improvement.

\subsubsection{Sensitivity analysis}\label{sec:sensitivity-method}

To assess the robustness of adapted parameters, for each parameter, we report the \emph{coefficient of variation}~(CV):
\begin{equation}\label{eq:cv}
\text{CV} = \frac{s}{\lvert\bar{v}\rvert},
\end{equation}
where $\bar{v} = \frac{1}{N}\sum_{i=1}^{N} v_i$ and $s = \sqrt{\frac{1}{N-1}\sum_{i=1}^{N}(v_i - \bar{v})^2}$ are the mean and standard deviation of the adapted values $\{v_i\}_{i=1}^{N}$ across $N = 30$~tunnels. CV measures how much a parameter changes from tunnel to tunnel: (i)~a high CV ($\geq 0.06$) identifies parameters that are \emph{tunnel-responsive}; (ii)~$\text{CV} \approx 0$ indicates \emph{baseline corrections} that take the same improved value on every tunnel regardless of geometry. We further examine \emph{cross-LLM consistency}: for parameters that always trigger adaptation, we compare the adapted values, tunnel counts, and tunnel categories produced by each LLM independently.

% ══════════════════════════════════════════════════════════════════════════
\section{Results}\label{sec:results}
% ══════════════════════════════════════════════════════════════════════════

By comparing segmentation performance against the static SAM4Tun baseline and a deterministic non-LLM control, we quantify the contributions of memory (m), state (s), and knowledge (k) across LLMs, identify consistently adapted parameters, and analyse residual errors and runtime trade-offs.

\subsection{Overall performance}\label{sec:main-results}

\begin{figure*}[t]
\centering
\includegraphics[width=0.8\textwidth]{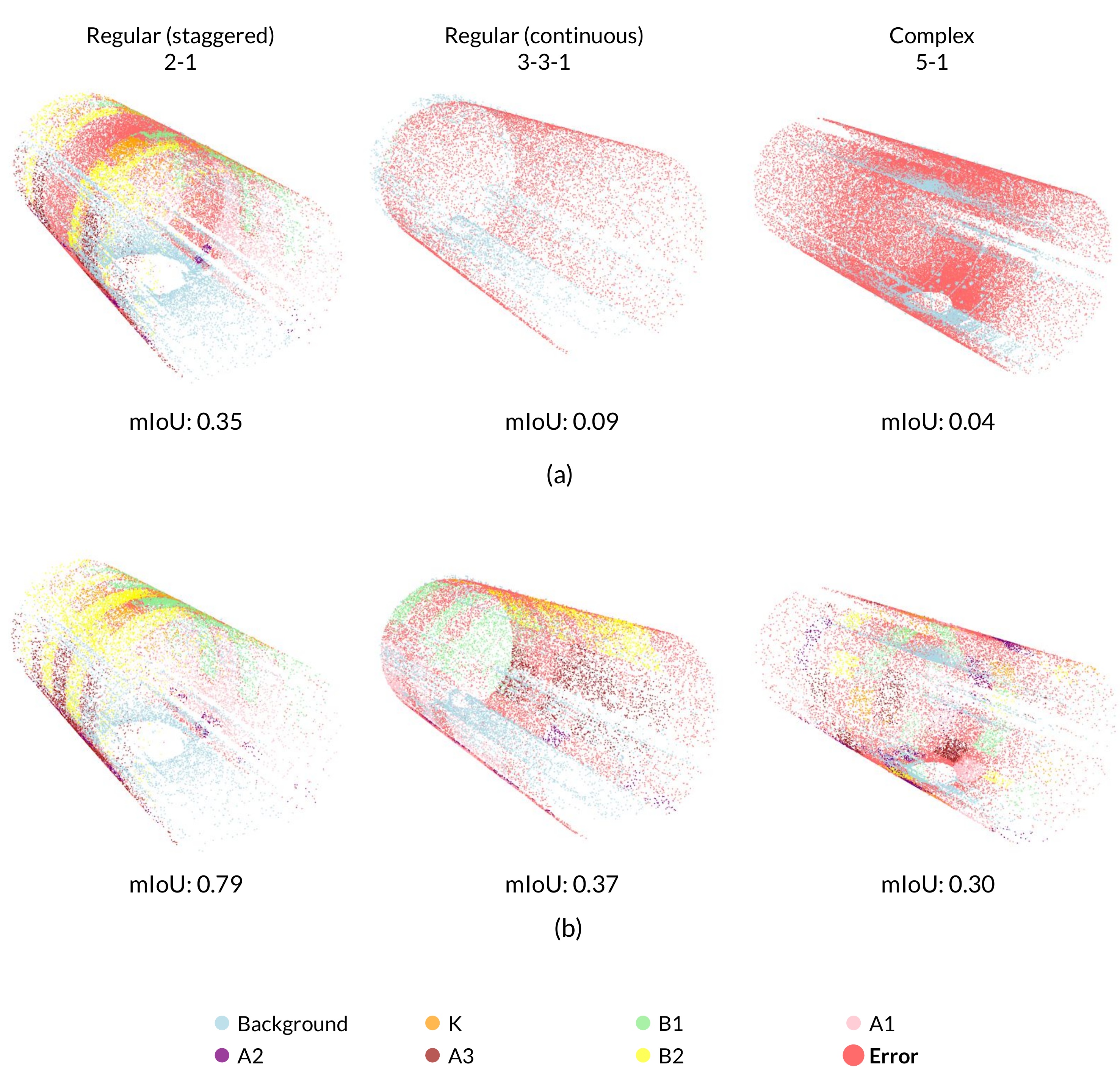}
\caption{Point-cloud visualisation for the tunnels with the largest mIoU gain in each category: regular (staggered; 2-1), regular (continuous; 3-3-1), and complex (5-1)}. Top (a): SAM4Tun baseline; bottom (b): LLM-adapted pipeline (m+s+k, Opus-4.6). Colours indicate classes; red denotes error; mIoU is shown per example.
\label{fig:ablation_point}
\end{figure*}

Before assessing adaptation across the 30 diverse subsets, we verified the corrected SAM4Tun implementation on the single curated tuning tunnel. Under the expert-tuned configuration, it reaches mIoU $\approx$ 0.88 (Table~\ref{tab:reference-tunnel}); under the same \mbox{$m+s+k$} adaptation, all three LLMs match or slightly surpass this expert anchor.

\begin{table}[ht]
\caption{Segmentation performance on the reference tunnel: expert configuration versus LLM-adapted (\mbox{$m+s+k$}).}\label{tab:reference-tunnel}
\begin{tabular*}{\tblwidth}{@{} l c c c c @{}}
\toprule
Metric & Expert & GPT-5.4 & Opus-4.6 & Gemini-3-Flash \\
\midrule
OA   & 0.9369 & 0.9390 & 0.9419 & 0.9354 \\
F1   & 0.9361 & 0.9389 & 0.9431 & 0.9362 \\
mIoU & 0.8801 & 0.8852 & 0.8926 & 0.8805 \\
\bottomrule
\end{tabular*}
\end{table}

Table~\ref{tab:main-results} summarises mIoU across conditions and LLMs. Under the full m+s+k design, overall mIoU rises from 0.18 (baseline) to 0.43--0.48 across the three LLMs ($p < 0.0001$, paired Cohen's \mbox{$d$} = 1.51--1.95); overall OA rises from 0.42 to 0.59--0.65. The improvement holds across both tunnel categories (Fig.~\ref{fig:ablation-bar}). Regular tunnels improve from 0.27 to 0.68--0.71 (paired Cohen's \mbox{$d$} = 1.4--2.2). Within the regular category, the near-reference staggered subsets (T1, T2), which match the reference tunnel in diameter, ring spacing, and key-segment layout, reach mIoU 0.784--0.796 across LLMs, whereas the continuous subset (T3), which departs from the reference in key-segment ordering, reaches only mIoU 0.270--0.309 and remains the main source of the lower combined regular figure. Complex tunnels, where the baseline drops to mIoU\,=\,0.04 because several pipeline assumptions no longer match the tunnel conditions, improve to 0.15--0.20 (paired Cohen's \mbox{$d$} = 1.2--2.5); complex-tunnel OA rises from 0.30 to 0.33--0.42 (Fig.~\ref{fig:ablation_point}).

Wall-clock runtime, API-call counts, and per-tunnel input/output token usage with the corresponding indicative USD cost for each LLM under m+s+k are reported in Appendix~\ref{app:practical} (Table~\ref{tab:practical}). Per-class IoU analysis (Appendix~\ref{app:perclass}) supports broad improvement across structural classes for regular tunnels and progressive recovery from near-zero baselines for complex tunnels. The full performance distribution is summarised in Appendix~\ref{app:distribution}.

\mbox{\textbf{Comparison against the non-LLM adaptation.}} The rule-based adaptation (Table~\ref{tab:main-results}, \emph{Non-LLM}) increases overall mIoU from 0.18 to 0.25, which shows some gains are deterministic.

On complex tunnels, the rules raise mIoU from 0.04 to 0.15 by applying the large-tunnel specification (diameter, ring spacing, segment count, key-segment layout). The LLM then adds a smaller lift to 0.15--0.20 (10/17 improved). The largest gap is on regular tunnels, where the LLM improve the mIoU from 0.27 to 0.68--0.71. Both non-LLM and LLM approaches remain low in absolute terms on complex tunnels under the single-reference setup.

\begin{figure*}[t]
\centering
\includegraphics[width=\textwidth]{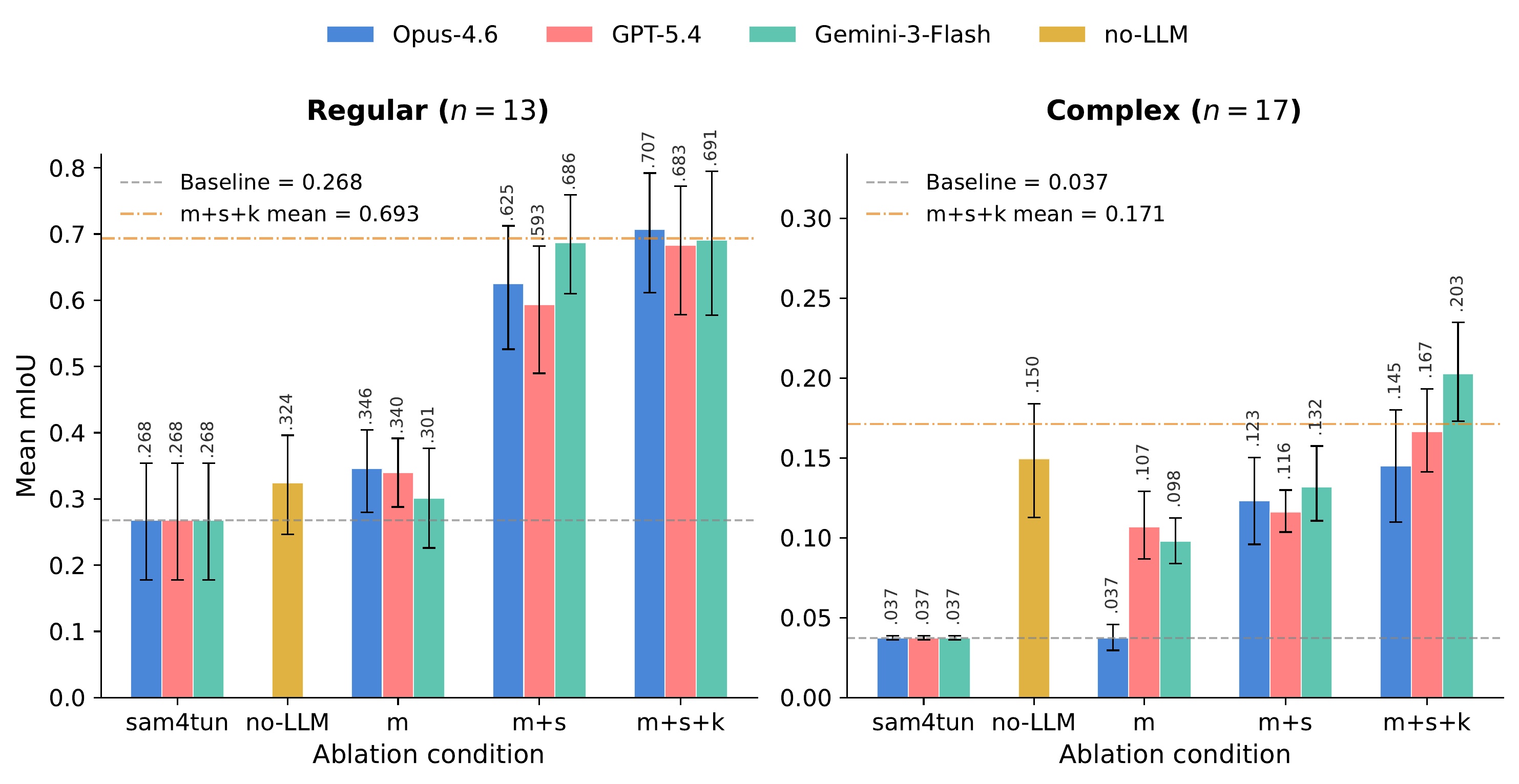}
\caption{Step-wise adaptation results split by tunnel categories.Bars left to right: \mbox{SAM4Tun} (static baseline), \mbox{\textit{Non-LLM}} (rule-based adaptation), \mbox{m}, \mbox{m+s}, \mbox{m+s+k}. The rule baseline closes part of the gap on complex tunnels but stays flat on regular tunnels; LLM (\mbox{m+s+k}) is highest in both categories. Error bars are bootstrap 95\% CIs.}
\label{fig:ablation-bar}
\end{figure*}

\begin{table*}[t]
\caption{Overall segmentation results ($n=30$): mIoU and paired $\Delta$mIoU vs. baseline with $p$-values, effect sizes, and bootstrap 95\% CIs. $\Delta$ is paired against \mbox{\texttt{SAM4Tun}}. OA is reported as a secondary metric, shown as the first line in each condition segment.}\label{tab:main-results}
\begin{tabular*}{\textwidth}{@{\extracolsep{\fill}} l l c c c c @{}}
\toprule
Condition & Statistic & Non-LLM & Opus-4.6 & GPT-5.4 & Gemini-3-Flash \\
\midrule
\multirow{2}{*}{Baseline (SAM4Tun)}
  & Mean OA & 0.42 & 0.42 & 0.42 & 0.42 \\
  & mIoU & 0.18 & 0.18 & 0.18 & 0.18 \\
\midrule
\multirow{7}{*}{Non-LLM}
  & Mean OA & 0.48 & --- & --- & --- \\
  & mIoU & 0.25 & --- & --- & --- \\
  & Mean $\Delta$mIoU & $+0.08$ & --- & --- & --- \\
  & $p$-value & 0.0035 & --- & --- & --- \\
  & paired Cohen's \mbox{$d$} & 0.58 & --- & --- & --- \\
  & std~($\Delta$) & 0.13 & --- & --- & --- \\
  & 95\% CI & $[0.028,\; 0.128]$ & --- & --- & --- \\
\midrule
\multirow{6}{*}{m}
  & Mean OA & --- & 0.46 & 0.48 & 0.43 \\
  & mIoU & --- & 0.22 & 0.25 & 0.22 \\
  & Mean $\Delta$mIoU & --- & $+0.05$ & $+0.07$ & $+0.04$ \\
  & $p$-value & --- & 0.558 & 0.028 & 0.056 \\
  & paired Cohen's \mbox{$d$} & --- & 0.11 & 0.42 & 0.36 \\
  & 95\% CI & --- & $[-0.113,\; 0.207]$ & $[0.008,\; 0.134]$ & $[-0.002,\; 0.090]$ \\
\midrule
\multirow{6}{*}{m+s}
  & Mean OA & --- & 0.60 & 0.58 & 0.63 \\
  & mIoU & --- & 0.42 & 0.40 & 0.47 \\
  & Mean $\Delta$mIoU & --- & $+0.25$ & $+0.23$ & $+0.29$ \\
  & $p$-value & --- & $<0.0001$ & $<0.0001$ & $<0.0001$ \\
  & paired Cohen's \mbox{$d$} & --- & 1.77 & 1.32 & 1.46 \\
  & 95\% CI & --- & $[0.196,\; 0.300]$ & $[0.163,\; 0.291]$ & $[0.215,\; 0.363]$ \\
\midrule
\multirow{6}{*}{m+s+k}
  & Mean OA & --- & 0.59 & 0.62 & 0.65 \\
  & \textbf{mIoU} & --- & \textbf{0.43} & \textbf{0.45} & \textbf{0.48} \\
  & Mean $\Delta$mIoU & --- & $+0.25$ & $+0.27$ & $+0.30$ \\
  & $p$-value & --- & $<0.0001$ & $<0.0001$ & $<0.0001$ \\
  & paired Cohen's \mbox{$d$} & --- & 1.95 & 1.95 & 1.51 \\
  & 95\% CI & --- & $[0.205,\; 0.301]$ & $[0.222,\; 0.327]$ & $[0.228,\; 0.378]$ \\
\bottomrule
\end{tabular*}
\end{table*}

\subsection{Ablation analysis}\label{sec:ablation}

To isolate each context component's contribution, Table~\ref{tab:cumulative_miou} reports cumulative mIoU gains at each ablation step. Fig.~\ref{fig:ablation-bar} visualises the step-wise progression for regular and complex categories across all three LLMs.

\begin{table}[ht]
\caption{Cumulative mIoU contribution (mean $\Delta$ vs. previous level).}\label{tab:cumulative_miou}
\begin{tabular*}{\tblwidth}{@{} l c c c @{}}
\toprule
Transition & Opus-4.6 & GPT-5.4 & Gemini-3-Flash \\
\midrule
Baseline $\rightarrow$ \mbox{\textit{Non-LLM}} & $+0.08$ & $+0.08$ & $+0.08$ \\
Baseline $\rightarrow$ m & $+0.05$ & $+0.07$ & $+0.04$ \\
\textbf{m $\rightarrow$ m+s} & $\mathbf{+0.20}$ & $\mathbf{+0.16}$ & $\mathbf{+0.25}$ \\
m+s $\rightarrow$ m+s+k & $+0.00$ & $+0.05$ & $+0.01$ \\
\bottomrule
\end{tabular*}
\end{table}
\textbf{Memory} alone provides an initial reference but is insufficient. On regular tunnels, especially the staggered subsets, memory alone changes mIoU by 0.04--0.07 across the three LLMs (Table~\ref{tab:cumulative_miou}); all three LLMs show small positive averages. Without intermediate feedback, the agent can propose parameter updates but cannot verify whether they improve or degrade the pipeline outputs.

\textbf{State} accounts for the largest observed increment in the ablation study. The m+s condition yields the strongest gain relative to baseline across all three LLMs (all $p < 0.0001$; paired Cohen's \mbox{$d$} = 1.32--1.77), while the memory-only effect is small or inconsistent, indicating that state is the main contributor to the observed improvement. State provides the agent with explicit quantitative evidence of how each stage has transformed the data: radial percentiles for mask bounds, retention rates for denoising aggressiveness, coverage uniformity for upsampling targets. Without state, the agent has only pre-pipeline statistics that may not reflect conditions after unfolding, denoising, or enhancing. The same conclusion holds when the comparator is the Non-LLM baseline rather than the static SAM4Tun: state alone (m+s) lifts overall mIoU by 0.16--0.25 above the rule-based adaptation.

One plausible explanation is that the gain associated with state depends on how the model maps raw numeric summaries (percentiles, counts, ratios) to parameter values in light of parameter semantics. We did not test alternative mapping strategies (e.g., regression models); however, the continuous, multidimensional nature of the characteristic space and the non-trivial parameter interactions suggest that capturing this mapping through static rules alone would require considerable manual engineering effort per tunnel category.

\textbf{Knowledge} provides a small additional gain, concentrated in the complex category. On top of m+s, knowledge raises mIoU by up to 0.05 across the three LLMs. The mean increments are small, but the per-tunnel direction is nevertheless positive. The increment is concentrated where the knowledge document supplies specific tuning guidance that neither memory nor state can reliably infer from numerical summaries alone (e.g.,  a larger-diameter tunnel often uses wider rings).

% (e.g., the coupling between ring diameter and spacing, where \texttt{ring\_spacing\_constant} should increase with diameter).

\subsection{Cross-model consistency and repeatability}\label{sec:consistency}

To assess whether the adaptation behaviour is model-dependent under a fixed prompt and input structure, Table~\ref{tab:cross-model} compares the three LLMs on the full m+s+k condition. The three models show similar effect ranges under the full m+s+k condition, although overlapping 95\% confidence intervals do not by themselves rule out between-model differences.
A repeatability check under the full \mbox{m+s+k} setting indicates low but non-zero stochastic drift. Across 30 tunnels, each LLM was inferred twice: on average, 90.9\% of the 18 critical parameters remain unchanged, and the mean absolute inter-run \mbox{$\Delta$}mIoU is 0.029 (median 0.000), as reported in Table~\ref{tab:repeatability} (Appendix~\ref{app:repeatability}).

\begin{table}[ht]
\caption{Cross-model summary (m+s+k vs baseline, overall).}\label{tab:cross-model}
\begin{tabular*}{\tblwidth}{@{} l c c c @{}}
\toprule
LLM & Mean $\Delta$mIoU & 95\% CI & \mbox{$d$} \\
\midrule
Opus-4.6 & $+0.25$ & $[0.21,\; 0.30]$ & 1.95 \\
GPT-5.4 & $+0.27$ & $[0.22,\; 0.33]$ & 1.95 \\
Gemini-3-Flash & $+0.30$ & $[0.23,\; 0.38]$ & 1.51 \\
\bottomrule
\end{tabular*}
\end{table}

\subsection{Runtime trade-off}\label{sec:runtime-tradeoff}

R4Tun trades additional API latency for bounded, logged parameter decisions. In this implementation, the SAM4Tun processing takes approximately 235\,s per tunnel, and the LLM adaptation under m+s+k adds 96--307\,s depending on the LLM, for total times of roughly 331--542\,s per tunnel (Table~\ref{tab:runtime-tradeoff}). The added runtime is therefore not negligible, but it is modest relative to the cost of annotating data or retraining a supervised model.

\begin{table}[ht]
\caption{Runtime trade-off of the tested R4Tun m+s+k setting.}\label{tab:runtime-tradeoff}
\begin{tabular*}{\tblwidth}{@{} l c c c @{}}
\toprule
LLM & Extra time & Total time & Mean $\Delta$mIoU \\
\midrule
Gemini-3-Flash & $+96$\,s ($0.4\times$) & $\sim$331\,s & $+0.30$ \\
Opus-4.6 & $+140$\,s ($0.6\times$) & $\sim$375\,s & $+0.25$ \\
GPT-5.4 & $+307$\,s ($1.3\times$) & $\sim$542\,s & $+0.27$ \\
\bottomrule
\end{tabular*}
\end{table}

\subsection{Parameter sensitivity}\label{sec:sensitivity}

This analysis separates parameters that vary across tunnels from those that remain constant, clarifying which aspects of the adaptation respond to specific tunnel conditions.

Analysis of 270 adapted parameter runs identified 18 parameters that all three LLMs consistently adjust (Table~\ref{tab:critical-params}). Of these, 11 are tunnel-responsive ($\text{CV} \geq 0.06$), in that their adapted values vary with tunnel geometry, clustering by tunnel category. The remaining 7 parameters behave as baseline corrections ($\text{CV} \approx 0$), with near-identical values across tunnels, indicating a shared correction trend relative to the SAM4Tun defaults.

\begin{table*}[t]
\caption{Critical parameters identified across all three LLMs (18 total). \emph{Tunnel-responsive} parameters ($\text{CV} \geq 0.06$) vary with tunnel geometry; \emph{baseline corrections} ($\text{CV} \approx 0$) take near-identical values across tunnels. The meaning/control column briefly explains what each parameter does in the SAM4Tun pipeline.}\label{tab:critical-params}
\begin{tabular*}{\textwidth}{@{\extracolsep{\fill}} l l p{0.23\textwidth} l c c l @{}}
\toprule
Stage & Parameter & Meaning/Control & Behaviour & Tunnels & CV & Adapted range / correction \\
\midrule
Unfolding & \texttt{diameter} & Cylindrical scale. & Responsive & 27/30 & 0.072 & [5.31, 7.6] \\
\midrule
Denoising & \texttt{mask\_r\_low} & Inner radial gate. & Responsive & 30/30 & 0.082 & [2.09, 3.75] \\
Denoising & \texttt{mask\_r\_high} & Outer radial gate. & Responsive & 30/30 & 0.147 & [2.78, 4.38] \\
Denoising & \texttt{default\_cutoff\_z} & Fallback radial cutoff (low density). & Responsive & 29/30 & 0.142 & [2.65, 6.27] \\
Denoising & \texttt{z\_step} & Radial histogram bin width. & Responsive & 30/30 & 0.181 & [0.003, 0.005] \\
Denoising & \texttt{smooth\_win} & Cutoff-curve smoothing window. & Correction & 30/30 & $\approx$0 & 3 $\rightarrow$ 5 \\
Denoising & \texttt{smooth\_offset} & Post-smoothing cutoff bias. & Correction & 30/30 & $\approx$0 & $-$0.003 $\rightarrow$ $-$0.002 \\
Denoising & \texttt{grad\_thresh} & Outer-edge gradient threshold. & Correction & 30/30 & $\approx$0 & 0.2 $\rightarrow$ 0.15 \\
Denoising & \texttt{y\_step} & Angular bin width (density filtering). & Correction & 30/30 & $\approx$0 & 0.5 $\rightarrow$ 0.4 \\
\midrule
Enhancing & \texttt{inter\_radius} & Interpolation neighbour radius. & Responsive & 30/30 & 0.130 & [0.03, 0.08] \\
Enhancing & \texttt{upsampling\_stage1} & Stage-1 upsampling spacing. & Responsive & 30/30 & 0.064 & [0.055, 0.11] \\
Enhancing & \texttt{curv\_thresh} & Panel--joint curvature threshold. & Correction & 30/30 & $\approx$0 & 0.0005 $\rightarrow$ 0.005 \\
Enhancing & \texttt{depth\_low} & Lower interpolation depth tolerance. & Correction & 30/30 & $\approx$0 & 0.003 $\rightarrow$ 0.005 \\
Enhancing & \texttt{depth\_high} & Upper interpolation depth tolerance. & Correction & 30/30 & $\approx$0 & 0.008 $\rightarrow$ 0.015 \\
\midrule
Segmenting & \texttt{hough\_thresh\_obliq} & Oblique joint-line vote threshold. & Responsive & 30/30 & 0.188 & [20, 83] \\
Segmenting & \texttt{hough\_thresh\_horiz} & Horizontal joint-line vote threshold. & Responsive & 30/30 & 0.204 & [20, 83] \\
Segmenting & \texttt{hough\_thresh\_vert} & Vertical ring-boundary vote threshold. & Responsive & 28/30 & 0.219 & [320, 980] \\
Segmenting & \texttt{processing.padding} & Segment-crop horizontal padding. & Responsive & 29/30 & 0.265 & [160, 419] \\
\bottomrule
\end{tabular*}
\end{table*}

\subsection{Error analysis}\label{sec:error-analysis}

\begin{figure*}[t]
\centering
\includegraphics[width=\textwidth]{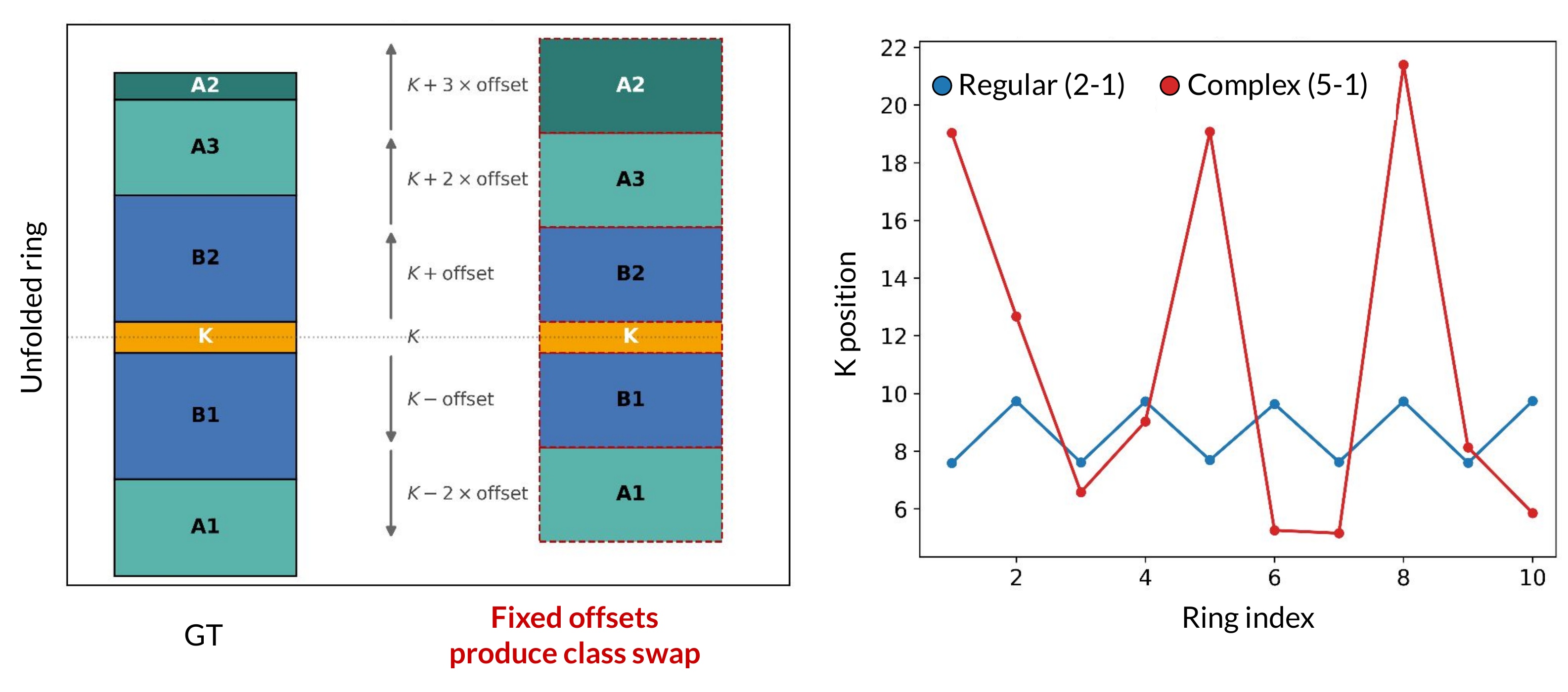}
\caption{Structural constraints of the fixed SAM4Tun labelling rule. (a)~Labelling rule: after K is detected, B segments are placed one step either side of K and A segments at successive multiples of the same fixed offset, in a fixed order, so the template mislabels rings whose segment widths or count differ from the reference. (b)~Ground-truth K position versus ring index for one regular (blue, 2-1) and one complex (red, 5-1) tunnel: K jumps from ring to ring, much more on the complex tunnel, which the current one-configuration-per-tunnel setting does not address.}
\label{fig:error-mechanism}
\end{figure*}

To understand why the absolute metrics stay low, we sorted every predicted point into four outcomes against the ground truth: \textit{correct}; \textit{false negative} (a segment point left as background, i.e. under-segmentation); \textit{false positive} (a background point pulled into a segment, i.e. over-segmentation, rare here); and \textit{class swap} (a correctly located segment point assigned to the wrong segment type).

\begin{table}[ht]
\caption{Error composition as a fraction of ground-truth points (Opus-4.6 \mbox{$m+s+k$} versus SAM4Tun baseline; 13 regular, 17 complex tunnels). FN: segment predicted as background; FP: background predicted as segment; Swap: segment predicted as the wrong segment class. Adaptation reduces false negatives; on complex tunnels the residual error is dominated by class swaps.}\label{tab:error-composition}
\begin{tabular*}{\tblwidth}{@{} l l c c c c @{}}
\toprule
Category & Method & Correct & FN & FP & Swap \\
\midrule
\multirow{2}{*}{Regular}
  & SAM4Tun       & 55\% & 22\% & 3\% & 20\% \\
  & \mbox{$m+s+k$} & 83\% & 6\%  & 4\% & 7\%  \\
\midrule
\multirow{2}{*}{Complex}
  & SAM4Tun       & 30\% & 70\% & 0\% & 0\%  \\
  & \mbox{$m+s+k$} & 41\% & 21\%  & 5\% & 32\% \\
\bottomrule
\end{tabular*}
\end{table}

Regular and complex tunnels fail differently (Table~\ref{tab:error-composition}). On regular tunnels, adaptation reduces under-segmentation, lowering FN from 22\% to 6\% of ground-truth points, and reduces class swaps from 20\% to 7\%. On complex tunnels, the baseline predicts almost all points as background; adaptation recovers the blocks (FN 70\%$\rightarrow$21\%). However, the fixed segment-offset template (anchored at K) turns a per-ring angular shift into systematic adjacent-class swaps (0\%$\rightarrow$32\%). Fig.~\ref{fig:error-regular-complex} shows this spatially: after adaptation, swaps dominate the residual error on complex tunnels.

\begin{figure*}[t]
\centering
\includegraphics[width=\textwidth]{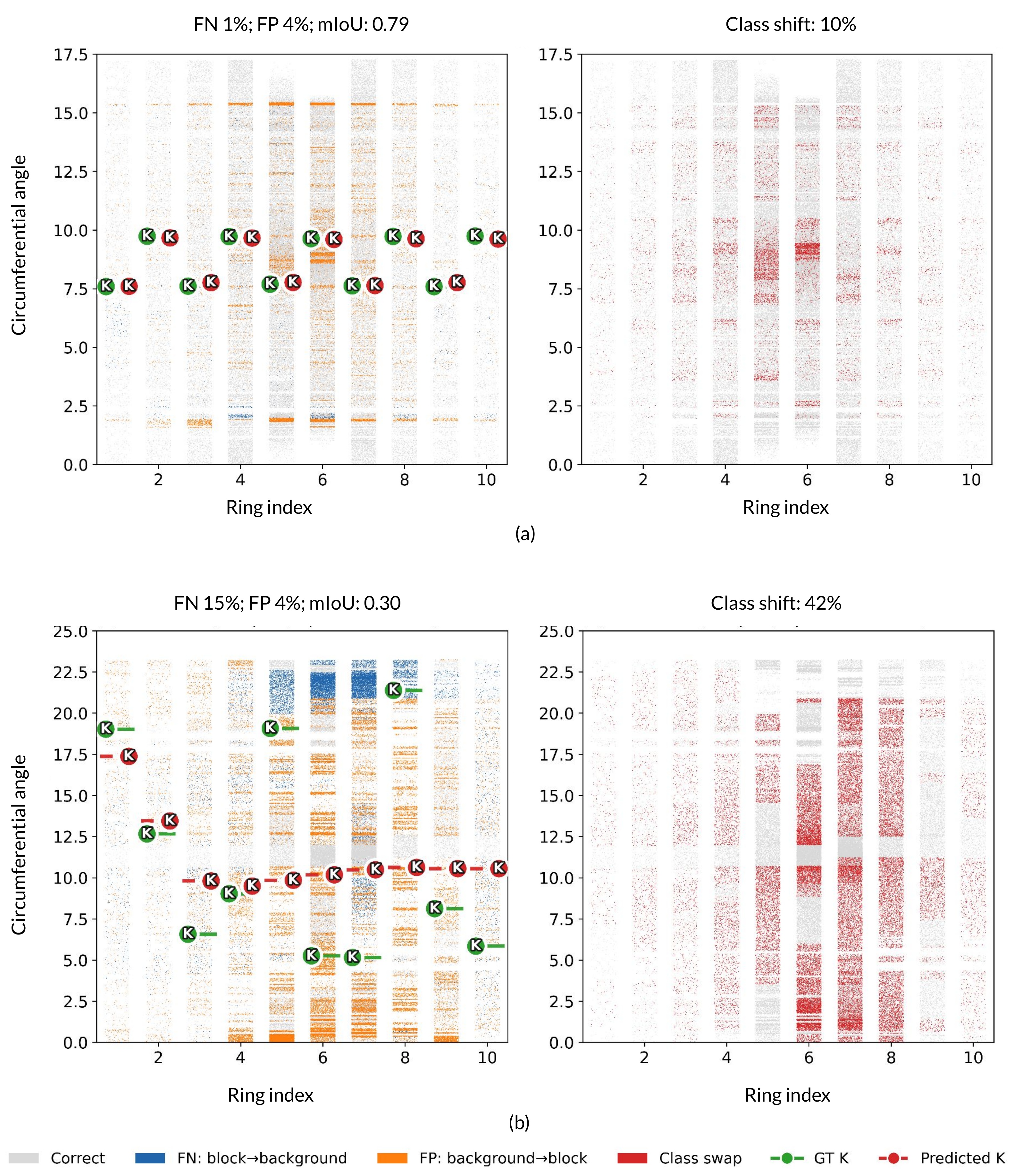}
\caption{Per-point error maps for a representative regular tunnel (a, regular tunnel 2-1) and complex tunnel (b, complex tunnel 5-1) under \mbox{$m+s+k$} (Opus-4.6), drawn on the unfolded surface (x: ring index; y: circumferential angle). \textbf{Left column}---detection outcomes: correct (grey), false negatives (blue, segment$\rightarrow$background) and false positives (orange, background$\rightarrow$segment), with the ground-truth K (green) and predicted K (red) overlaid per ring; titles give FN, FP, and mIoU. \textbf{Right column}---class swaps (red) among otherwise correctly located points. On the regular tunnel the predicted K stays on the GT K and swaps remain low (10\%). On the complex tunnel the predicted K is nearly constant and drifts away from the GT K, which jumps ring-to-ring; each affected ring's labels then rotate together and swaps rise to 42\%. The maps localise the two constraints: the K mismatch (left) drives the moving-anchor errors, and the resulting label rotation (right) is the dominant residual error.}
\label{fig:error-regular-complex}
\end{figure*}

The failure modes trace back to a mismatch between how the current open-source SAM4Tun pipeline is parameterised and how segmental rings actually vary. The pipeline applies \emph{one parameter set per tunnel} and labels each ring by first detecting the key segment (K) and then placing every other segment at a \emph{fixed} angular offset from K, in a \emph{fixed} order. Neither of these two structural constraints can be removed by tunnel-level parameter tuning.

\begin{enumerate}
\item \emph{Per-ring density variation under tunnel-level parameterisation.} R4Tun adapts one configuration per tunnel, but point density is far from uniform \emph{within} a tunnel: the per-ring point count varies by up to 19-fold (regular) and 38-fold (complex) between the dense central rings and the sparse end rings. A single tunnel-level setting cannot keep both ends of this range inside the detector's working band, so the sparse rings systematically lose ring-boundary and K detection no matter how the tunnel-level parameters are chosen. This is a limitation of the adaptation \emph{scope} (one configuration per tunnel), not of any individual parameter value, and it can only be lifted by moving from tunnel-level to ring-level adaptation.
\item \emph{Fixed segment-offset template.} Once K is located, the rule places the B segments one step on either side of K and the A segments at successive multiples of that same fixed step, in a fixed order (Fig.~\ref{fig:error-mechanism}a). The step size and the segment order are hard-coded, so when the true geometry departs from the reference the fixed steps no longer line up with the real joints and correctly detected points are still given the wrong label. This is the single largest source of class swaps, visible as the red bands in the right column of Fig.~\ref{fig:error-regular-complex}, which grow from 10\% (regular) to 42\% (complex) even where points are correctly located.
\end{enumerate}

Lifting these limits is future work rather than a parameter change, because they are baked into SAM4Tun's one-configuration-per-tunnel, detect-K-then-fixed-template logic. The clearest direction is ring-level adaptation: density-adaptive preprocessing and per-ring K re-anchoring would address the density and K-detection problems above; a variable-length segment template and handedness inference would relax the fixed labelling rule. Implementing this requires rewriting the open-source pipeline and lies outside the bounded-parameter adaptation studied here. These directions extend SAM4Tun and R4Tun rather than invalidate them: structured context already enables LLM adaptation within fixed bounds, and pinpoints which structural parts must change next to lift the absolute ceiling.

% ══════════════════════════════════════════════════════════════════════════
\section{Discussion}\label{sec:discussion}
% ══════════════════════════════════════════════════════════════════════════

Taken together, the results support a mechanism-level claim rather than a deployment-level one. With the SAM4Tun pipeline and the expert reference held fixed, structured context improves bounded parameter adaptation across three LLMs. In practical terms, \mbox{$m+s+k$} can guide a reviewable parameter update without labelled retraining; however, at current accuracy it supports assisted analysis, not autonomous inspection.

\subsection{Key findings}

The experimental evaluation yields two primary findings. First, as noted in the ablation study (Section~\ref{sec:ablation}), the ablation results suggest that the context design improves adaptation across models and tunnel categories, with state providing the main gains ($+0.16$ to $+0.25$ mIoU on top of memory). Second, the LLM (m+s+k) gain over the non-LLM rule-based adaptation is concentrated on the regular category, where the rule table cannot improve over the static baseline (Section~\ref{sec:main-results}). This concentration is consistent with the single-reference design: both comparators start from a configuration tuned on a tunnel that is similar to the regular category. Near the reference, the LLM has more informative context to condition on, and its advantage over deterministic lookup becomes visible; far from the reference (the complex category), neither approach has a matching anchor and the two converge toward similarly low absolute performance, with the LLMs retaining a small mean advantage. As reported in Section~\ref{sec:main-results}, the regular category gap occurs where the deterministic control does not improve over the static baseline, whereas the complex category convergence more likely reflects a shared ceiling than a specific failure of LLM reasoning (Section~\ref{sec:limitations}).

In practice, a domain expert calibrates the pipeline on a single reference tunnel, encodes the tuning rationale into structured documents, and deploys R4Tun for subsequent tunnels. Under this single-reference setup, the framework is best suited to targets that remain close to the calibrated reference, and it can also flag when the fixed pipeline begins to fail. The framework supports post-hoc review by logging a rationale alongside each parameter change, and requires no model retraining or labelled domain data. Experts must still define the reference configuration and bounds, review rationales, and judge whether low-confidence outputs are acceptable. Within this scope, R4Tun does not yet support autonomous final inspection or downstream tasks (e.g. structural health monitoring or Scan-to-BIM) without human verification.

% ── requires amssymb for \checkmark; cas-dc usually loads it. If not, add:
% \usepackage{amssymb}
\subsection{Comparison with state-of-the-art}\label{sec:positioning}

We position R4Tun against established tunnel-segmentation paradigms in Table~\ref{tab:sota}. The comparison is presented as an operating-point and setup-cost comparison rather than a direct accuracy ranking, because the methods are not evaluated under comparable conditions. Supervised 3D deep-learning models on Seg2Tunnel (LiningNet, SparseUNet, SCF-Net, FA-ResNet) report mIoU 0.83--0.90~\cite{lin2024seg2tunnel}, but are evaluated in-distribution and require per-point labels and GPU retraining. Geometric feature-engineering methods avoid training but depend on manually tuned thresholds and typically report cross-section fitting error rather than per-segment mIoU, so they are not directly comparable at the component level. Expert-tuned SAM4Tun reaches mIoU above 0.90, but only on curated, per-case-tuned test rings~\cite{ye2025sam4tun}; under our protocol (30 diverse subsets, single fixed reference, no per-case tuning) the corrected open-source pipeline reaches mIoU 0.88 on its own reference tunnel but only 0.18 when that same fixed configuration is applied static across all subsets.

The only directly comparable pair (same evaluation protocol, with no labels, no training, and no per-case tuning) is R4Tun (\mbox{0.43--0.48}) versus static SAM4Tun (0.18); R4Tun improves the directly comparable baseline. With the corrected implementation and anchor, R4Tun also reaches mIoU 0.784--0.796 across LLMs on near-reference (regular-staggered) tunnels. Supervised and expert-tuned figures come from more favourable settings and bound the remaining gap, not direct competitors. Deterministic, non-LLM rule baseline (Section~\ref{sec:main-results}) reaches only 0.25, and every higher figure in Table~\ref{tab:sota} requires labels, training, or per-case expert tuning.

\begin{table*}[t]
\caption{Positioning R4Tun against representative tunnel-segmentation paradigms. The comparison is on setup cost as well as accuracy: \checkmark{} indicates the requirement (or capability) applies, ``--'' that it does not. mIoU figures are \emph{not} measured under comparable settings (see the ``setting'' column): supervised and expert-tuned figures are taken from~\cite{lin2024seg2tunnel} and~\cite{ye2025sam4tun} under their original, more favourable protocols, whereas the static-SAM4Tun, non-LLM, and R4Tun figures are from this study under a single fixed reference across 30 diverse subsets. R4Tun is the only approach that is simultaneously label-free, training-free, and expert-tuning-free while still adapting per tunnel.}\label{tab:sota}
\begin{tabular*}{\textwidth}{@{\extracolsep{\fill}} l c c c c l @{}}
\toprule
Method & Labels & Training & \begin{tabular}[t]{@{}c@{}}Per-case\\expert tuning\end{tabular} & \begin{tabular}[t]{@{}c@{}}Per-tunnel\\adaptation\end{tabular} & mIoU (setting) \\
\midrule
LiningNet/SparseUNet/SCF-Net/FA-ResNet & \checkmark & \checkmark & -- & -- & 0.83--0.90 (in-distribution) \\
Expert-tuned SAM4Tun & -- & -- & \checkmark & -- & above 0.90 (curated rings) \\
Static SAM4Tun (this protocol) & -- & -- & -- & -- & 0.18 \\
Non-LLM rules (this study) & -- & -- & -- & \checkmark & 0.25 \\
\textbf{R4Tun (near-reference)} & -- & -- & -- & \checkmark~ & \textbf{0.784--0.796 (across LLMs)} \\
R4Tun (all tunnels) & -- & -- & -- & \checkmark~ & 0.43--0.48 \\
\bottomrule
\end{tabular*}
\end{table*}

\subsection{Limitations}\label{sec:limitations}
The limitations of this work are driven by the fixed SAM4Tun operator design and by the single-reference adaptation protocol. In particular, R4Tun inherits SAM4Tun's assumptions and failure modes, so the LLM can only re-parameterise the pipeline rather than replace it.
All parameter adaptation is anchored to a single expert-tuned configuration, which creates a low ceiling for both the LLM-guided and the rule-based methods. When no contextual anchor resembles the target tunnel, neither approach has sufficient information to recover strong performance. Expanding R4Tun to encode multiple expert-tuned reference configurations (e.g., one per tunnel category) may further narrow the absolute gap.

Our study was evaluated exclusively on the SAM4Tun pipeline and the Seg2Tunnel dataset; potentially, its transferability is most plausible when four conditions hold: (i) each stage exposes bounded, tunable parameters (i.e., adaptation is a re-parameterisation problem rather than a redesign); (ii) expert reference configurations exist whose target characterisation is comparable to the new deployment setting; (iii) each stage admits a compact, summarised intermediate state that can be passed between agents; and (iv) stage-specific knowledge can be encoded (e.g., as bounds, rules, and failure signatures) for use during adaptation.

% ══════════════════════════════════════════════════════════════════════════
\section{Conclusions}\label{sec:conclusions}
% ══════════════════════════════════════════════════════════════════════════

In this paper, we presented R4Tun, positioned as a mechanism contribution rather than a deployable final-inspection system, a multi-agent framework that extends a fixed, expert-designed tunnel-lining segmentation pipeline (SAM4Tun) with LLM-guided bounded parameter adaptation. Rather than replacing the underlying geometric operators, each stage agent compares a new tunnel against the reference configuration, reads compact summaries of intermediate pipeline outputs, and proposes bounded parameter updates accompanied by a logged rationale, preserving the deterministic pipeline structure while supporting post-hoc expert review. We validated R4Tun on 30 Seg2Tunnel subsets (13 regular, 17 complex) across three different LLMs. The main findings of this study are:

\begin{itemize}
\item Structured context improves bounded adaptation across LLMs. The full \mbox{$m+s+k$} design raised mIoU from 0.18 to 0.43--0.48 and OA from 0.42 to 0.59--0.65; on the near-reference regular (staggered) subsets it reached mIoU 0.784--0.796, approaching the expert anchor (0.88).
\item Intermediate pipeline state is the dominant contributor. Adding state increased mIoU by 0.16--0.25 over memory alone, whereas memory or knowledge in isolation produced only small increments.
\item The adaptation behaviour is consistent across models. The three LLMs produced overlapping effect ranges, adjusted a shared set of 18 critical parameters under identical prompts, and showed limited run-to-run drift (90.9\% of critical parameters unchanged across repeats).
\item The framework is label-free and auditable. R4Tun shifts expert effort from repeated per-tunnel intervention toward an upfront authoring step (reference calibration plus per-stage knowledge documents), runs on commercial LLM APIs without labelled retraining, logs a rationale for each parameter change, and adds only modest runtime (96--307\,s per tunnel).
\end{itemize}

These results also reveal limitations that are inherent to the present design. First, all adaptation is anchored to a single expert-tuned reference, which sets a low ceiling away from that anchor: the regular-continuous (T3) and complex subsets differ from the reference in segment arrangement, segment count, and diameter, and remain challenging (mIoU 0.15--0.31), so at current accuracy R4Tun supports assisted, expert-verified parameter re-configuration for near-reference tunnels rather than autonomous final inspection or unverified downstream use such as structural health monitoring or Scan-to-BIM. Second, R4Tun adapts but does not replace SAM4Tun's operators; in particular, its one-configuration-per-tunnel parameterisation and fixed offset-and-order labelling template turn per-ring geometric variation into systematic class swaps that persist regardless of parameter choice (\Cref{sec:error-analysis}). Finally, all evidence is confined to the SAM4Tun--Seg2Tunnel setting, and transferability to other parameter-controllable pipelines is argued conceptually (\Cref{sec:limitations}) but not yet empirically validated.

In the future, encoding multiple expert-tuned reference anchors (for example, one per tunnel category) would narrow the single-anchor gap for off-reference tunnels. Replacing the fixed-template labelling logic with dynamic, ring-level adaptatio would lift the structural ceiling identified in the error analysis. Validating the \mbox{$m+s+k$} strategy on other parameter-controllable backbones would test its expected backbone-agnostic behaviour, and broader validation across tunnel typologies and acquisition conditions, together with human-in-the-loop integration into downstream inspection tasks, would strengthen generalisability and move the framework toward verified deployment.

% ══════════════════════════════════════════════════════════════════════════
% Data availability, declarations, etc.
% ══════════════════════════════════════════════════════════════════════════

\section*{Data availability}

The source code, adapted parameters, and evaluation scripts are available at \url{https://github.com/Tao-Robominds/R4Tun}. The Seg2Tunnel dataset is publicly available.

\section*{Declaration of competing interest}

The authors declare that they have no known competing financial interests or personal relationships that could have appeared to influence the work reported in this paper.

\section*{Declaration of generative AI use}

During this study, the authors evaluated large language models (Opus-4.6, GPT-5.4, and Gemini-3-Flash) as the main experimental systems. The LLMs were also used for language editing and for improving manuscript structure. The authors reviewed and edited all content and take full responsibility for the content of the publication.

%\printcredits  % Uncomment if cas-dc supports it in your version

% ══════════════════════════════════════════════════════════════════════════
% Bibliography
% ══════════════════════════════════════════════════════════════════════════

% \clearpage % start Biography on a new page
% \section*{Biography}
% TODO: add biography text here

% Natbib requires a BibTeX style; without it, citations stay as “?”
% NOTE: \balance can cause column-balancing artefacts in the reference list (e.g.,
% lines slipping into the footer area) with this class. Keep it off for the bibliography.
%\balance
\bibliographystyle{model1-num-names}
\bibliography{references}

% \input{output.bbl}

% ══════════════════════════════════════════════════════════════════════════
% Appendices (included after the references)
% ══════════════════════════════════════════════════════════════════════════

\clearpage
% \section*{Appendices}

% ══════════════════════════════════════════════════════════════════════════
% Appendices — previously included via \input{appendices}
% Ordered to follow the paper structure:
%   A  Baseline parameter tables          (§3.1–3.3, §3.5 sam4tun)
%   B  Characteriser fields               (§3.4 R4Tun overview)
%   C  Context components example         (§3.4.2 Context design)
%   D  Worked CoT trace                   (§3.4.3 CoT design)
%   E  On-site rules baseline             (§3.5 baselines)
%   F  Runtime, API calls, and cost       (§3.6 implementation)
%   G  Per-class IoU breakdown            (§4 results)
%   H  Performance distribution           (§4 results)
% ══════════════════════════════════════════════════════════════════════════

\FloatBarrier
% Use numeric appendix section numbers (1, 2, 3, ...) instead of letters (A, B, C, ...).
% (The default \appendix switches \thesection to \Alph{section}.)
\setcounter{section}{0}
\renewcommand{\thesection}{\arabic{section}}
\renewcommand{\thesubsection}{\thesection.\arabic{subsection}}

% ──────────────────────────────────────────────────────────────────────────
\section{Baseline parameter tables}\label{app:params}
Tables~\ref{tab:unfold-params}--\ref{tab:segment-params} report the SAM4Tun baseline parameter values used as the reference configuration for all adaptation experiments.

\begin{table}[ht]
\caption{Stage~1 --- Unfolding parameters (sam4tun baseline).}\label{tab:unfold-params}
\begin{tabular*}{\tblwidth}{@{} l c @{}}
\toprule
Parameter & Value \\
\midrule
delta & 0.005 \\
slice\_spacing\_factor & 1.2 \\
vertical\_filter\_window & 4.5 \\
ransac\_threshold & 1 \\
ransac\_probability & 0.9 \\
ransac\_inlier\_ratio & 0.75 \\
ransac\_sample\_size & 5 \\
polynomial\_degree & 3 \\
num\_samples\_factor & 1210 \\
diameter & 5.60 \\
\bottomrule
\end{tabular*}
\end{table}

\begin{table}[ht]
\caption{Stage~2 --- Denoising parameters (sam4tun baseline).}\label{tab:denoise-params}
\begin{tabular*}{\tblwidth}{@{} l c @{}}
\toprule
Parameter & Value \\
\midrule
mask\_r\_low & 2.7 \\
mask\_r\_high & 2.8 \\
y\_step & 0.5 \\
z\_step & 0.001 \\
grad\_threshold & 0.2 \\
smoothing\_window\_size & 3 \\
smoothing\_offset & $-0.003$ \\
default\_cutoff\_z & 2.7 \\
\bottomrule
\end{tabular*}
\end{table}

\begin{table}[ht]
\caption{Stage~3 --- Enhancing parameters (sam4tun baseline).}\label{tab:enhance-params}
\begin{tabular*}{\tblwidth}{@{} l c @{}}
\toprule
Parameter & Value \\
\midrule
upsamp\_stage1\_target\_dist & 0.08 \\
upsamp\_stage2\_target\_dist & 0.04 \\
upsamp\_stage3\_target\_dist & 0.02 \\
curvature\_threshold & 0.0005 \\
depth\_threshold\_low & 0.003 \\
depth\_threshold\_high & 0.01 \\
inter\_radius & 0.06 \\
duplicate\_threshold & 0.02 \\
num\_neighbors & 20 \\
num\_interpolations & 2 \\
resolution & 0.005 \\
window\_size & 9 \\
\bottomrule
\end{tabular*}
\end{table}

\begin{table}[ht]
\caption{Stage~4 --- Segmenting parameters (sam4tun baseline). \textsuperscript{*}}\label{tab:segment-params}
\begin{tabular*}{\tblwidth}{@{} l c @{}}
\toprule
Parameter & Value \\
\midrule
\multicolumn{2}{@{}l}{\textit{Boundary detection}} \\
binary\_threshold & 127 \\
morph\_kernel\_size & [3, 3] \\
dilation\_iterations & 1 \\
hough\_thresh\_oblique & 50 \\
minLineLength\_oblique & 100 \\
maxLineGap\_oblique & 40 \\
hough\_thresh\_horiz & 50 \\
minLineLength\_horiz & 100 \\
maxLineGap\_horiz & 10 \\
hough\_thresh\_vert & 500 \\
angle\_range\_obliq\_pos & [6, 9] \\
angle\_range\_obliq\_neg & [$-9$, $-6$] \\
merge\_distance & 3 \\
ring\_spacing\_constant & 1.2 \\
resolution & 0.005 \\
\midrule
\multicolumn{2}{@{}l}{\textit{SAM template}} \\
segment\_per\_ring & 6 \\
segment\_order & [K, B1, A1, A2, A3, B2] \\
segment\_width & 1200 \\
K\_height & 1079.92 \\
AB\_height & 3239.77 \\
angle & 7.52 \\
processing.padding & 150 \\
processing.y\_bounds & [4200, 13100] \\
processing.crop\_margin & 50 \\
\bottomrule
\end{tabular*}
\end{table}

% ──────────────────────────────────────────────────────────────────────────
\section{Characteriser fields}\label{app:chars}
Table~\ref{tab:chars} summarises the characteriser fields used to describe each tunnel and to populate the per-stage state provided to the agents.

\begin{table}[ht]
\caption{Characteriser fields by stage. Raw fields are available to all stages; each subsequent group becomes available after the corresponding stage completes.}\label{tab:chars}
\begin{tabular*}{\columnwidth}{@{\extracolsep{\fill}} l p{0.72\columnwidth} @{}}
\toprule
Source & Fields \\
\midrule
\multirow{3}{*}{Raw} & estimated diameter, tunnel length, tunnel height \\
 & $z$-range (min, max) \\
 & mean / median / min nearest-neighbour distance \\
\midrule
\multirow{4}{*}{Unfolded} & $r$-percentiles ($p_{10}$, $p_{99}$), $h$-span, $\theta$-span, $\theta$-range \\
 & median / std nearest-neighbour distance \\
 & intensity median, intensity min \\
\midrule
\multirow{3}{*}{Denoised} & mean / median / std nearest-neighbour distance \\
 & estimated diameter, tunnel length, surface completeness \\
 & surface regularity, average curvature, section curvatures \\
\midrule
\multirow{3}{*}{Enhanced} & total points, median / mean nearest-neighbour distance \\
 & coverage uniformity \\
 & template spacing suitability, current median spacing \\
\bottomrule
\end{tabular*}
\end{table}

% ──────────────────────────────────────────────────────────────────────────
\section{Non-LLM rule-based pseudocode}\label{app:rules}\label{sec:rule-baseline}

The non-LLM baseline reads the same per-stage knowledge documents the LLM agents read, transcribed into a deterministic Python lookup. The parameter-selection logic for the denoising stage is reproduced below. Other stages follow the same pattern.

% The verbatim block must stay together; if there isn't enough space left in
% the current column/page, force a break before it (without flushing floats).
\pagebreak[4]
\begin{verbatim}
def select_denoising(chars):
    diameter = chars["estimated_diameter"]
    density  = chars["density"]
    p10      = chars["unfolded_p10"]
    p99      = chars["unfolded_p99"]

    if diameter > 6.5:
        family = "large"          
    elif chars["joint_type"] == "continuous":
        family = "continuous"     
    else:
        family = "base"           

    params = REFERENCE_DENOISING.copy()

    if family == "large":
        params["mask_r_low"]       = p10
        params["mask_r_high"]      = p99 + 0.05
        params["smooth_win"]       = 5
    elif family == "continuous":
        params["mask_r_high"]      = max(p99, 2.85)
        params["smooth_win"]       = 6
    return params
\end{verbatim}

% ──────────────────────────────────────────────────────────────────────────
\section{Context components: denoising agent example}\label{app:context}

This appendix reproduces the three context components (memory, state, knowledge) that the agent receives as input.

\subsection{Memory: reference vs target raw characteristics}\label{app:context-memory}
Memory pairs the reference tunnel's raw characteristics with those of the target tunnel using an identical schema, so the agent can compute deviations field-by-field (Table~\ref{tab:context-memory}).

\begin{table}[ht]
\caption{Memory excerpt (raw characteristics). Reference is the expert-tuned tunnel; target is tunnel~4-1.}\label{tab:context-memory}
\begin{tabular*}{\tblwidth}{@{} l c c c @{}}
\toprule
Field & Reference & Target (4-1) & $\Delta$ \\
\midrule
Total points & 1{,}109{,}768 & 1{,}872{,}537 & $+69\%$ \\
Estimated diameter (m) & 5.32 & 7.41 & $+39\%$ \\
Tunnel length (m) & 12.16 & 18.10 & $+49\%$ \\
Tunnel height (m) & 5.08 & 7.72 & $+52\%$ \\
Mean NN distance (m) & 0.0082 & 0.0081 & $-1\%$ \\
Median NN distance (m) & 0.0065 & 0.0065 & $\approx 0\%$ \\
\bottomrule
\end{tabular*}
\end{table}

Memory also bundles the SAM4Tun reference parameters for the stage (Table~\ref{tab:denoise-params}), so the agent always has a known-good baseline to deviate from.

\subsection{State: cumulative outputs of upstream stages}\label{app:context-state}
State grows as the pipeline executes. For the denoising agent, state consists of the unfolded characteristics produced by the preceding unfolding stage, supplied as a reference/target pair (Table~\ref{tab:context-state}).

\begin{table}[ht]
\caption{State excerpt (unfolded characteristics, after the unfolding stage).}\label{tab:context-state}
\begin{tabular*}{\tblwidth}{@{} l c c @{}}
\toprule
Field & Reference & Target (4-1) \\
\midrule
$r$-percentile $p_{10}$ (m) & 2.30 & 2.38 \\
$r$-percentile $p_{99}$ (m) & 2.77 & 3.93 \\
$h$-span (m) & 12.08 & 18.86 \\
$\theta$-span (rad) & 17.28 & 23.28 \\
Median NN distance (m) & 0.047 & 0.066 \\
\bottomrule
\end{tabular*}
\end{table}

\subsection{Knowledge: parameter semantics, ranges, and constraints}\label{app:context-knowledge}
Knowledge is a stage-specific Markdown document, authored once and shared across all tunnels. It enumerates each parameter together with its empirically validated range, defaults, and inter-parameter constraints. The denoising knowledge document is reproduced below.

\begin{quote}\small
\textbf{Cross-tunnel variation.}
Tunnel lining datasets commonly vary along several axes:
\textit{tunnel scale}, from smaller metro-scale tunnels to larger-diameter tunnels;
\textit{ring geometry}, including shorter or longer ring lengths;
\textit{segment layout}, including different numbers and ordering of lining segments per ring;
\textit{joint assembly}, such as staggered, continuous, or interleaved joint arrangements;
and \textit{scanning configuration}, including single-station scans, multi-station registration, or uneven scanner placement.

\textbf{Tunable parameters.}
\texttt{mask\_r\_low}~(m), inner radial gate before depth histogramming, range $[2.09, 3.75]$ (baseline 2.7);
\texttt{mask\_r\_high}~(m), outer radial gate, range $[2.78, 4.38]$ (baseline 2.8);
\texttt{default\_cutoff\_z}~(m), fallback radial cutoff when a $\theta$-bin lacks reliable counts, range $[2.65, 6.27]$ (baseline 2.7);
\texttt{z\_step}~(m), radial bin width per histogram column, range $[0.003, 0.005]$ (baseline 0.001).

\textbf{Proven defaults.}\par
\texttt{smoothing\_window\_size}\,=\,5,\par
\texttt{smoothing\_offset}\,=\,$-0.002$,\par
\texttt{grad\_threshold}\,=\,0.15,\par
\texttt{y\_step}\,=\,0.4.

\textbf{Diagnostic rules.}
\texttt{mask\_r\_low}\,$<$\,\texttt{mask\_r\_high};
\texttt{default\_cutoff\_z} should stay coherent with the upstream unfolding \texttt{diameter} (order of magnitude $\approx$ diameter/2, with diameter in metres).
\end{quote}

The three other stages (unfolding, enhancing, segmenting) follow the same four-block template; the per-stage knowledge documents are released in the project's code repository.

% ──────────────────────────────────────────────────────────────────────────
\section{Worked CoT trace}\label{app:cot}

The following trace continues the worked example from Appendix~\ref{app:context}: the denoising agent receives the memory, state, and knowledge inputs above, executes the five-step CoT protocol (Section~\ref{sec:cot}), and emits a single schema-conformant JSON object that the denoising stage consumes unchanged.

\paragraph{Step~1 --- Referencing.}
Estimated diameter is $7.41$\,m vs reference $5.32$\,m ($+39\%$). Unfolded $r$-percentiles are $p_{10}=2.38$, $p_{99}=3.93$, vs reference $p_{10}=2.30$, $p_{99}=2.77$. Median nearest-neighbour distance after unfolding is $0.066$\,m vs reference $0.047$\,m ($+41\%$), indicating lower areal density on the tunnel wall.

\paragraph{Step~2 --- Diagnostic inspection.}
Large-diameter, moderate-sparse target. The reference \texttt{mask\_r\_low}/\texttt{mask\_r\_high} ($2.7/2.8$) would clip all wall points beyond $r=2.8$\,m, which is below the target $p_{10}=2.38$ and well below $p_{99}=3.93$; the radial gate must widen substantially. \texttt{default\_cutoff\_z} (2.7) is below the target $p_{99}$ (3.93) and would force the fallback path to discard valid wall points; it must rise. Lower density ($+41\%$ NN distance) requires \texttt{z\_step} to relax from $0.001$ so each radial bin retains enough samples for the gradient test.

\paragraph{Step~3 --- Parameter adaptation.}
Set \texttt{mask\_r\_low}\,$=2.37$ (just below target $p_{10}=2.38$); \texttt{mask\_r\_high}\,$=3.80$ (just below target $p_{99}=3.93$, retaining $\geq 99\%$ of wall points while staying within the empirical range $[2.78, 4.38]$); \texttt{default\_cutoff\_z}\,$=3.7$ (close to \texttt{mask\_r\_high} and order $\approx$ diameter/2 $= 3.71$); \texttt{z\_step}\,$=0.005$ (proven default for sparse scans). All other parameters take the proven defaults from the knowledge block.

\paragraph{Step~4 --- Validation.}
\texttt{mask\_r\_low}\,(2.37) $<$ \texttt{mask\_r\_high}\,(3.80); both lie within their empirical ranges. \texttt{default\_cutoff\_z}\,(3.7) is consistent with the unfolding \texttt{diameter}\,(7.41\,m, half-diameter $3.71$\,m). \texttt{z\_step}\,(0.005) lies in $[0.003, 0.005]$. All keys present, all types match Table~\ref{tab:denoise-params}.

\paragraph{Step~5 --- JSON output.}
\begin{verbatim}
    {
      "mask_r_low": 2.37,
      "mask_r_high": 3.80,
      "y_step": 0.4,
      "z_step": 0.005,
      "grad_threshold": 0.15,
      "smoothing_window_size": 5,
      "smoothing_offset": -0.002,
      "default_cutoff_z": 3.7
    }
\end{verbatim}

The same parameter keys are adapted by GPT-5.4 and Gemini~3~Flash in qualitatively the same direction for this tunnel.

% ──────────────────────────────────────────────────────────────────────────
\section{Runtime, API calls, and cost}\label{app:practical}
Table~\ref{tab:practical} reports the compute setup and, for one \texttt{m+s+k} run, the per-tunnel input/output token counts and indicative USD cost for each LLM (averaged over the 30 tunnels, four stage calls per tunnel).

\begin{table}[ht]
\caption{Runtime, API calls, and per-tunnel API cost. Token counts are as reported by each vendor API; USD figures use vendor list prices at submission and are indicative.}\label{tab:practical}
\begin{tabular*}{\tblwidth}{@{} l l @{}}
\toprule
Metric & Value \\
\midrule
LLM API calls per tunnel & 4 (one per adapted stage agent) \\
Total API calls (full ablation) & 120 per condition per LLM \\
GPU & Single NVIDIA RTX~5060 \\
Retraining required & None \\
Labelled data required & None (GT for evaluation only) \\
\midrule
\multicolumn{2}{@{}l}{\textit{Per-tunnel tokens and cost (m+s+k, mean over 30 tunnels)}} \\
Opus 4.6 & 64{,}854 in / 6{,}599 out, \$1.47 \\
GPT-5.4 & 64{,}414 in / 35{,}263 out, \$0.43 \\
Gemini 3 Flash & 83{,}658 in / 3{,}058 out, \$0.03 \\
\bottomrule
\end{tabular*}
\end{table}

% ──────────────────────────────────────────────────────────────────────────
\section{Repeatability analysis}\label{app:repeatability}
Table~\ref{tab:repeatability} reports a basic repeatability check for the full \mbox{m+s+k} condition (temperature~0). Each tunnel was re-inferred twice per LLM and the resulting adapted parameters and mIoU were compared between run~1 and run~2.

\begin{table}[ht]
\caption{Repeatability under \mbox{m+s+k} (temperature~0). Metrics are computed over 30 tunnels, with two runs per tunnel per LLM.}\label{tab:repeatability}
\begin{tabular*}{\tblwidth}{@{} l c @{}}
\toprule
Metric & Value \\
\midrule
Mean critical parameters unchanged (18 total) & 90.9\% \\
Mean $\lvert\Delta\mathrm{mIoU}\rvert$ (run~1 vs run~2) & 0.029 \\
Median $\lvert\Delta\mathrm{mIoU}\rvert$ (run~1 vs run~2) & 0.000 \\
\bottomrule
\end{tabular*}
\end{table}

% ──────────────────────────────────────────────────────────────────────────
% Place any pending floats before continuing, but avoid creating float-only blank pages.
\FloatBarrier
\pagebreak[4]

\section{Per-class IoU breakdown}\label{app:perclass}

Tables~\ref{tab:perclass-regular} and~\ref{tab:perclass-complex} report per-class IoU for Opus~4.6 (representative; other LLMs show the same pattern) alongside the rules baseline. For regular tunnels, rules achieve per-class IoU comparable to sam4tun while all LLM conditions improve roughly uniformly. For complex tunnels (rules $n=17$ including 3 failed tunnels scored as zero), rules recover segment structure from near-zero for most classes but not K-block; the LLM conditions achieve further improvement on regular tunnels. B2-block remains the hardest class, requiring the 7-segment layout guidance from the knowledge component.

\begin{table}[ht]
\caption{Per-class IoU---Regular tunnels ($n=13$, 6-class, Opus~4.6).}\label{tab:perclass-regular}
\begin{tabular*}{\tblwidth}{@{} l c c c c c @{}}
\toprule
Class & sam4tun & rules & memory & m+s & m+s+k \\
\midrule
Background & 0.511 & 0.639 & 0.576 & 0.671 & 0.698 \\
K-block & 0.151 & 0.295 & 0.372 & 0.599 & 0.558 \\
B1-block & 0.217 & 0.256 & 0.338 & 0.643 & 0.613 \\
A1-block & 0.268 & 0.234 & 0.317 & 0.681 & 0.651 \\
A2-block & 0.198 & 0.159 & 0.180 & 0.403 & 0.398 \\
A3-block & 0.302 & 0.272 & 0.320 & 0.667 & 0.679 \\
B2-block & 0.229 & 0.412 & 0.317 & 0.710 & 0.723 \\
\bottomrule
\end{tabular*}
\end{table}

\begin{table}[ht]
\caption{Per-class IoU---Complex tunnels ($n=17$, 7-class, Opus~4.6). Rules include 3 failed tunnels scored as zero.}\label{tab:perclass-complex}
\begin{tabular*}{\tblwidth}{@{} l c c c c c @{}}
\toprule
Class & sam4tun & rules & memory & m+s & m+s+k \\
\midrule
Background & 0.337 & 0.534 & 0.358 & 0.513 & 0.520 \\
K-block & 0.000 & 0.000 & 0.034 & 0.183 & 0.159 \\
B1-block & 0.000 & 0.041 & 0.001 & 0.084 & 0.135 \\
A1-block & 0.000 & 0.072 & 0.005 & 0.128 & 0.155 \\
A2-block & 0.000 & 0.083 & 0.006 & 0.143 & 0.116 \\
A3-block & 0.000 & 0.149 & 0.017 & 0.092 & 0.119 \\
A4-block & 0.000 & 0.116 & 0.019 & 0.100 & 0.104 \\
B2-block & 0.000 & 0.099 & 0.000 & 0.000 & 0.043 \\
\bottomrule
\end{tabular*}
\end{table}

% ──────────────────────────────────────────────────────────────────────────
\section{Performance distribution}\label{app:distribution}
Table~\ref{tab:distribution} summarises the distribution of mIoU across tunnels (reported as the mean across the three LLMs for each condition).

\begin{table}[ht]
\caption{Performance distribution (mean across 3 LLMs).}\label{tab:distribution}
\begin{tabular*}{\tblwidth}{@{} l c c c c c @{}}
\toprule
Metric & sam4tun & rules & memory & m+s & m+s+k \\
\midrule
Mean mIoU & 0.176 & 0.254 & 0.230 & 0.430 & 0.452 \\
Std & 0.173 & 0.165 & 0.135 & 0.304 & 0.282 \\
Min & 0.037 & 0.119 & 0.068 & 0.099 & 0.156 \\
Max & 0.451 & 0.562 & 0.397 & 0.829 & 0.791 \\
\bottomrule
\end{tabular*}
\end{table}

% Ensure the final appendix tables are placed (and references resolve).
\clearpage
\FloatBarrier

\end{document}